\documentclass{template}

\newcommand{\templateoption}{option3}

\usepackage[T1]{fontenc}
\usepackage{lmodern}  
\usepackage{xspace}
\usepackage{booktabs}
\usepackage{makecell}
\usepackage{wrapfig}
\usepackage{graphicx,amsmath,amssymb,hyperref}
\usepackage[authoryear,round]{natbib}
\usepackage[table]{xcolor}
\usepackage[percent]{overpic}
\usepackage{url}

\hypersetup{
  colorlinks=true,
  urlcolor=magenta,
  linkcolor=black,
  citecolor=link,
  filecolor=black,
  pdfborder={0 0 0}
}

\definecolor{link}{HTML}{0063BE}

\providecommand{\sectionname}{Section}
\providecommand{\appendixname}{Appendix}

\newcommand{\figref}[1]{%
  \figurename~\hyperref[#1]{\textcolor{link}{\ref*{#1}}}%
}
\newcommand{\tabref}[1]{%
  \tablename~\hyperref[#1]{\textcolor{link}{\ref*{#1}}}%
}
\newcommand{\secref}[1]{%
  \sectionname~\hyperref[#1]{\textcolor{link}{\ref*{#1}}}%
}
\newcommand{\appendixref}[1]{%
  \appendixname~\hyperref[#1]{\textcolor{link}{\ref*{#1}}}%
}

\newcommand{\tablestyle}[2]{\setlength{\tabcolsep}{#1}\renewcommand{\arraystretch}{#2}\centering\footnotesize}
\renewcommand{\paragraph}[1]{\textbf{#1.}}
\newcolumntype{x}[1]{>{\centering\arraybackslash}p{#1pt}}
\newcolumntype{y}[1]{>{\raggedright\arraybackslash}p{#1pt}}
\newcolumntype{z}[1]{>{\raggedleft\arraybackslash}p{#1pt}}
\setlist{nosep} 
\newcommand{\eg}{\emph{e.g}.}
\newcommand{\ie}{\emph{i.e}.}

\usepackage{subcaption}
\usepackage{float}

\newcommand{\vs}{\emph{vs}.\ }
\usepackage{multirow}
\newlength\savewidth\newcommand\shline{\noalign{\global\savewidth\arrayrulewidth
  \global\arrayrulewidth 1pt}\hline\noalign{\global\arrayrulewidth\savewidth}}
  
\newcommand{\benchmarkname}{UEval}
\makeatletter
\fancypagestyle{firststyle}{
  \fancyhead[L]{}
  \fancyhead[C]{}
  \fancyhead[R]{}
  \fancyfoot[L]{\raisebox{3pt}{\fontsize{9}{10}\selectfont \the\correspondingauthor}}
  \fancyfoot[R]{}
}
\makeatother

\makeatletter
\DeclareRobustCommand\bfseries{%
  \not@math@alphabet\bfseries\mathbf
  \fontseries\bfdefault\selectfont
  \sffamily
}
\makeatother

\DeclareTextFontCommand{\textbf}{\bfseries\sffamily}

\title{\centering \fontsize{18}{24}\selectfont{UEval: A Benchmark for Unified Multimodal Generation}}

\author{
    \vspace{.2cm}
    \parbox{\textwidth}{\centering
        Bo Li \quad
        Yida Yin \quad
        Wenhao Chai \quad
        Xingyu Fu$^{*}$ \quad
        Zhuang Liu$^{*}$
    }
    \\
    \vspace{-0.1cm}
    {\normalfont\fontsize{11}{15}\selectfont {Princeton University}}
     \vspace{0.2cm} \\ 
    $\vcenter{\hbox{\includegraphics[height=1.2em]{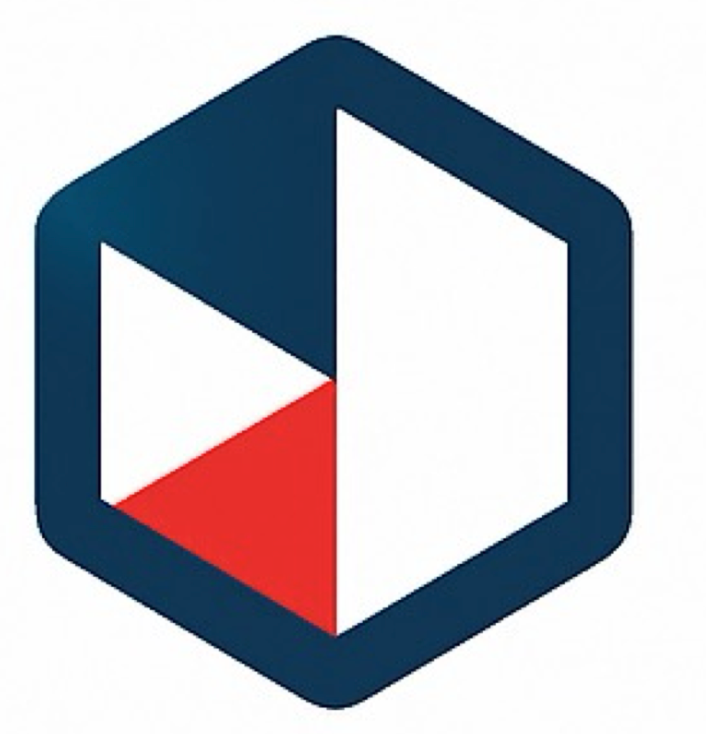}}}$\hspace{0.3em}
    \texttt{Website:} \url{https://zlab-princeton.github.io/UEval}
    \vspace{0.2cm} \\ 
    \raisebox{-0.4ex}{\includegraphics[height=1em]{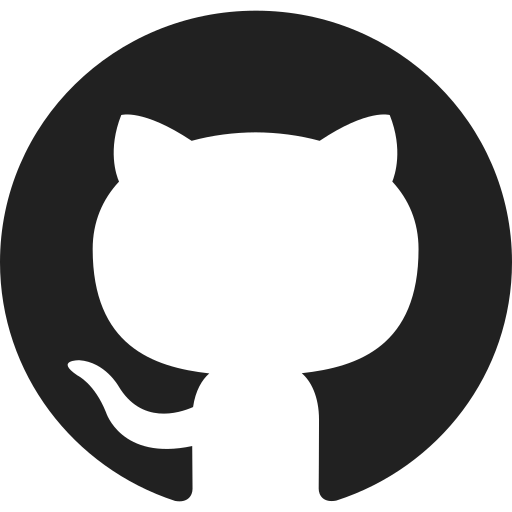}}\hspace{0.3em}\href{https://github.com/zlab-princeton/UEval_benchmark}{\texttt{Code}} 
    \hspace{0.2cm}
    \raisebox{-0.4ex}{\includegraphics[height=1em]{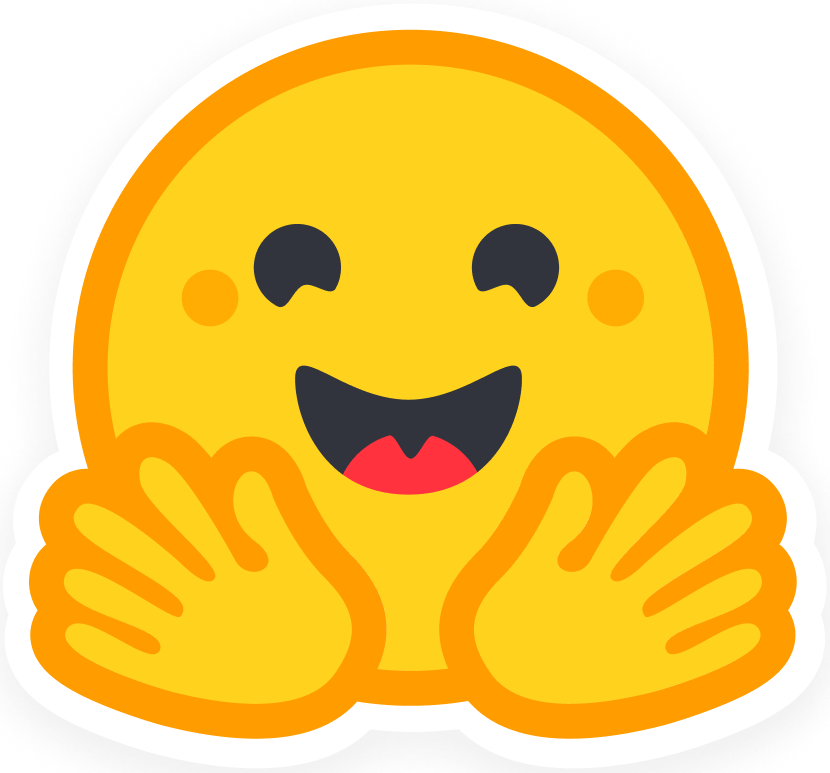}}\hspace{0.3em}\href{https://huggingface.co/datasets/zlab-princeton/UEval}{\texttt{Dataset}}
    \hspace{0.2cm}
    \raisebox{-0.4ex}{\includegraphics[height=1em]{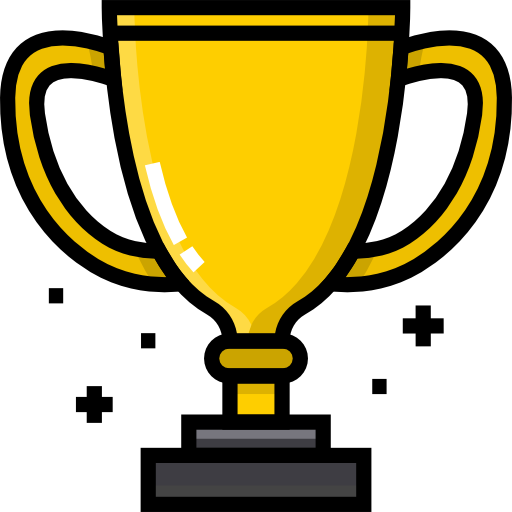}}\hspace{0.3em}\href{https://zlab-princeton.github.io/UEval/leaderboard/Full-Leaderboard/1.0}{\texttt{Leaderboard}}
    \vspace{-3mm}
}

\correspondingauthor{\, $^*$ Co-advising}

\makeatletter
\renewcommand{\abscontent}{}%
\makeatother

\newenvironment{abstractblock}{%
  {\centering\large\bfseries\sffamily Abstract\par}
  \vspace{0.2em}
  \begin{list}{}{%
      \setlength{\leftmargin}{2em}
      \setlength{\rightmargin}{2em}
      \setlength{\topsep}{0pt}
      \setlength{\parsep}{0pt}
  }
  \item[]
}{%
  \end{list}
  \par\normalfont\vspace{1em}
}

\begin{document}

\begingroup
\makeatletter
\let\raggedright\centering
\makeatother

\maketitle
\endgroup

\newcommand{\abstractcontent}{%
We introduce \benchmarkname, a benchmark to evaluate \emph{unified models}, \ie, models capable of generating both images and text. \benchmarkname{} comprises 1,000 expert-curated questions that require both images and text in the model output, sourced from 8 real-world tasks. Our curated questions cover a wide range of reasoning types, from step-by-step guides to textbook explanations. Evaluating open-ended multimodal generation is non-trivial, as simple LLM-as-a-judge methods can miss the subtleties.  
Different from previous works that rely on multimodal Large Language Models (MLLMs) to rate image quality or text accuracy, we design a rubric-based scoring system in \benchmarkname{}.  For each question, reference images and text answers are provided to a MLLM to generate an initial rubric, consisting of multiple evaluation criteria, and human experts then refine and validate these rubrics. In total, \benchmarkname{} contains 10,417 validated rubric criteria, enabling scalable and fine-grained automatic scoring.
\benchmarkname{} is challenging for current unified models: GPT-5-Thinking scores only 66.4 out of 100, while the best open-source model reaches merely 49.1. We observe that reasoning models often outperform non-reasoning ones, and transferring reasoning traces from a reasoning model to a non-reasoning model significantly narrows the gap. This suggests that reasoning may be important for tasks requiring complex multimodal understanding and generation. 
}

\newcommand{\teaserfigure}[2]{%
    \begin{figure}[#1]
        \centering
        \includegraphics[width=#2]{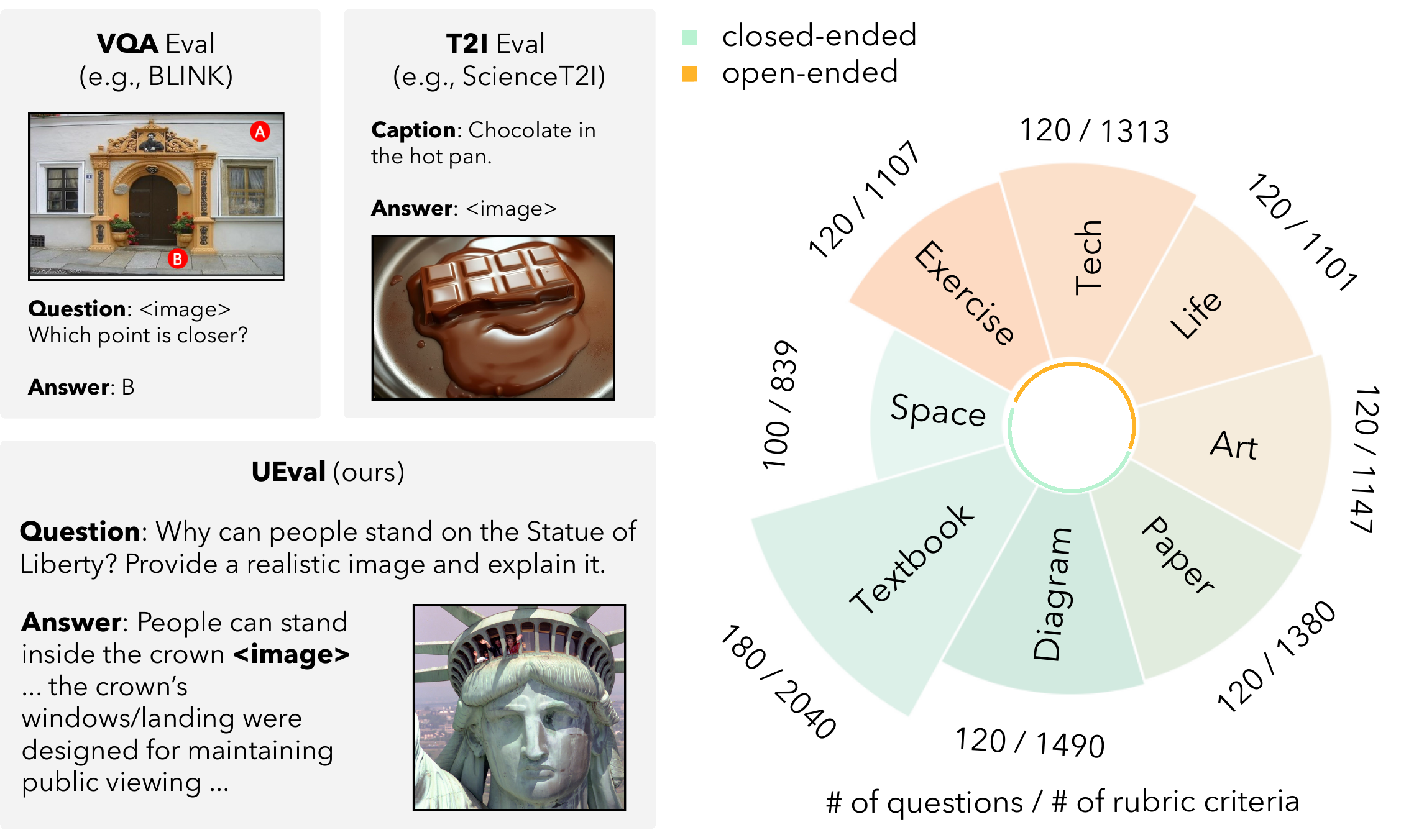}
        \caption{\textbf{Put here the caption for your figure.} Place your figure caption in this section. Your figure description goes here. Insert the caption for your figure here. Add caption here. Include the description of your figure here. Place additional details about the figure content in this area. Describe the visual elements shown in the figure.}
        \label{fig:teaser}
    \end{figure}%
}

\newcommand{\linkstablestyleone}{%
    \begin{center}
        \small
        \renewcommand{\arraystretch}{1.2}
        \begin{tabular}{rll}
            \worldwideweb & \textbf{Website} & \url{https://www.test.com/}\\
            \github & \textbf{Code} & \url{https://github.com/test}\\
            \hf & \textbf{Data} & \url{https://huggingface.com}
        \end{tabular}
    \end{center}%
}

\newcommand{\linkstablestyletwo}{%
    \noindent\small
    \textbf{Website:} \url{https://www.test.com/}\\
    \textbf{Code:}    \url{https://github.com/test}\\
    \textbf{Data:}    \url{https://huggingface.com}
}


\makeatletter
\ifthenelse{\equal{\templateoption}{option1}}{
    \vspace{-0.1cm}
    
    \teaserfigure{!ht}{\textwidth}

    \vspace{0.3cm}

    \begin{abstractblock}
    \abstractcontent
    \end{abstractblock}

    \newpage
}{
    \ifthenelse{\equal{\templateoption}{option2}}{
        \vspace{-0.3cm}
        
        \begin{abstractblock}
        \abstractcontent
        \end{abstractblock}

        \vspace{0.0cm}

        \teaserfigure{!bh}{\textwidth}

        \newpage
    }{
        \begin{abstractblock}
        \abstractcontent
        \end{abstractblock}

        \vspace{0.5cm}


        \vspace{-0.5cm}

    }
}
\makeatother

\section{Introduction}

Unified multimodal models~\citep{tong2024metamorph, zhou2024transfusionpredicttokendiffuse, deng2025emerging} aim to integrate multimodal understanding and generation capabilities within a single system. Current evaluations of these models are largely confined to two paradigms: visual question answering~\citep{marino2019okvqavisualquestionanswering, liu2024mmbench, yue2024mmmumassivemultidisciplinemultimodal,fu2024mmecomprehensiveevaluationbenchmark}, which requires generating a textual answer from one or more input images, and text-to-image generation~\citep{huang2023t2i, ghosh2023geneval, lin2024evaluating}, which takes a textual description as input and asks the model to produce a corresponding image. 

 These paradigms overlook a central component of multimodal reasoning scenarios: unified multimodal generation that \emph{produces both text and images} in response to a single query (\figref{fig:teaser}). In many real-world tasks, effective responses require images to illustrate specific concepts while simultaneously producing text to explain those visual elements. Without such evaluation, existing benchmarks fail to capture the rich interplay between language and vision that characterizes real-world multimodal reasoning.

While recent efforts~\citep{an2024openleaf, liu2024holistic, xia2024mmie, niu2025wise,zhao2025envisioning} have proposed new benchmarks to evaluate unified models, there remains a lack of standardized approaches for evaluating unified multimodal generation. To address this gap, we introduce \benchmarkname{}, a challenging benchmark to assess unified models~\citep{wang2024emu3, chen2025multimodal,yang2025mmada,nanobanana,xie2025show} at scale. Unlike prior benchmarks, \benchmarkname{} requires models to reason and respond to complex user queries jointly in images and natural language, providing a rigorous testbed across diverse real-world scenarios. 

\begin{figure}[t]
\centering
\includegraphics[width=\linewidth]{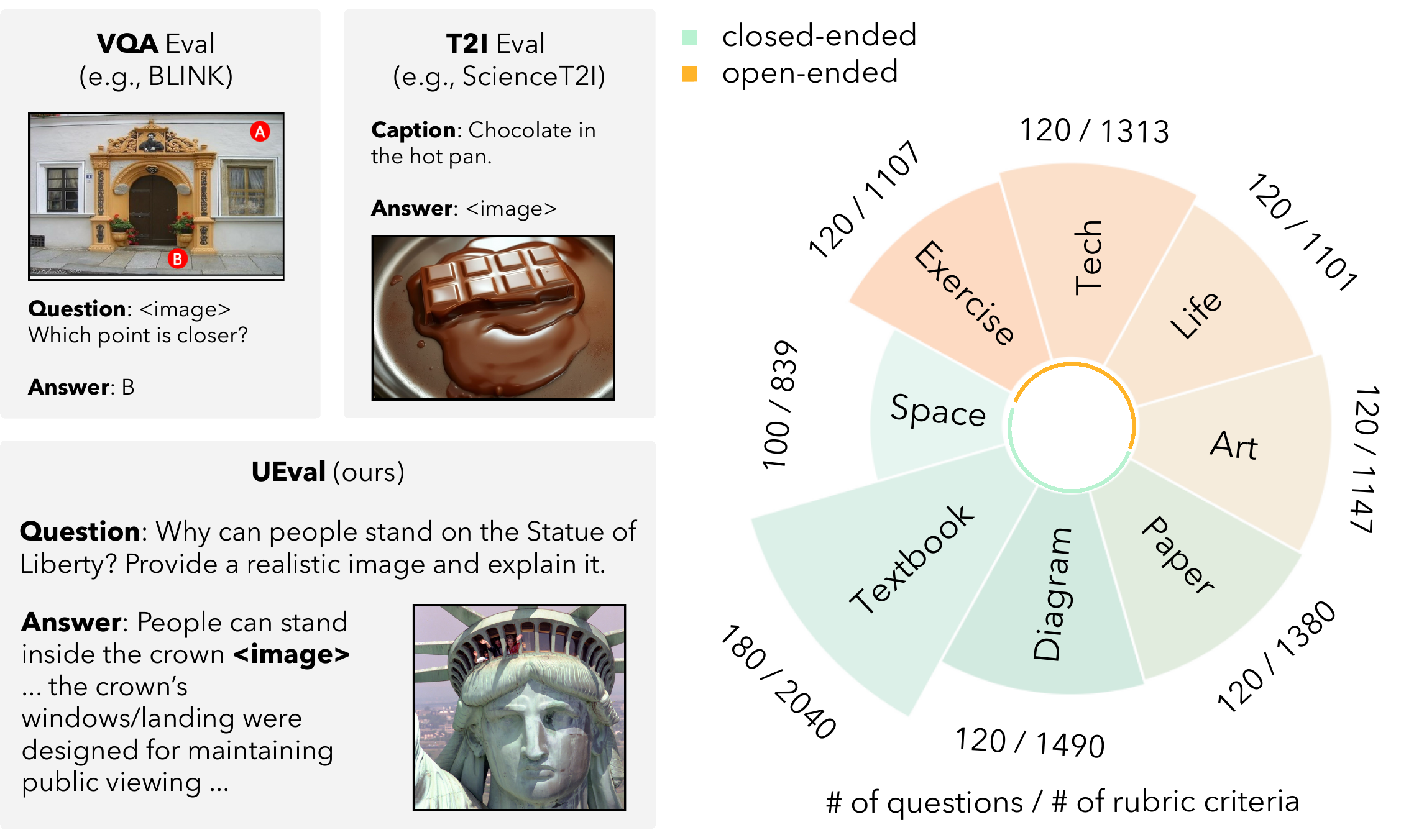}
\caption{\textbf{Left}: Previous unified model evaluations focus on either image understanding (\ie, VQA) or image generation from captions (\ie, T2I). In contrast, \benchmarkname{} requires models to reason across modalities and generate responses in both images and text. The VQA example is from BLINK~\citep{fu2024blink}, and the T2I example is from ScienceT2I~\citep{li2025science}. \textbf{Right}: The chart illustrates the number of questions and rubric criteria across tasks in \benchmarkname{}.}
\label{fig:teaser}
\end{figure}

\benchmarkname{} comprises 1,000 expert-curated questions spanning 8 diverse real-world tasks, including \textit{space}, \textit{textbook}, \textit{paper}, \textit{diagram}, \textit{art}, \textit{life}, \textit{tech}, and \textit{exercise}. Inspired by \citet{arora2025healthbenchevaluatinglargelanguage}, we propose a rubric-based framework for consistent and reproducible evaluation. 
For each question, we first manually collect reference answers in both text and image. Then, a frontier multimodal Large Language Model (MLLM) generates an initial rubric, consisting of multiple rubric criteria, conditioned on the original question and reference answers.
Human annotators further refine these rubrics to eliminate redundancies and add any missing criteria. In total, \benchmarkname{} contains 10,417 rigorously validated rubric criteria to enable reliable automatic grading. In our experiments, we employ Gemini-2.5-Pro~\citep{comanici2025gemini25pushingfrontier} as a judge model to score model responses with our rubrics and find that its scores show strong agreement with human judgments.

We conduct a comprehensive evaluation of 9 unified models on \benchmarkname{}. 
Our results show that \benchmarkname{} presents a challenge to all models. Among them, GPT-5-Thinking~\citep{gpt5} achieves the highest score of 66.4 out of 100 when averaged across tasks, whereas the best-performing open-source model (\ie, Emu3.5~\citep{cui2025emu35nativemultimodalmodels}) reaches only 49.1. We also observe that current models struggle to generate multiple images with consistent labeling across steps in multi-step planning tasks (\eg, drawing a cat step by step).

Interestingly, we observe that reasoning models (\eg, GPT-5-Thinking) outperform their non-reasoning variants (\eg, GPT-5-Instant) on most tasks. To further investigate the benefit of reasoning in multimodal generation, we append the reasoning trace produced by GPT-5-Thinking to the end of the original question prompt and feed it into non-reasoning models. 
Surprisingly, this substantially improves the visual outputs generated by GPT-5-Instant and Gemini-2.5-Flash, while open-source models (\eg, BAGEL) show no improvement. These observations suggest that Chain-of-Thought reasoning~\citep{wei2023chainofthoughtpromptingelicitsreasoning}, long studied in Large Language Models (LLMs), may also play an important role in unified multimodal generation.

\begin{figure}[t]
  \centering
  \includegraphics[width=\linewidth]{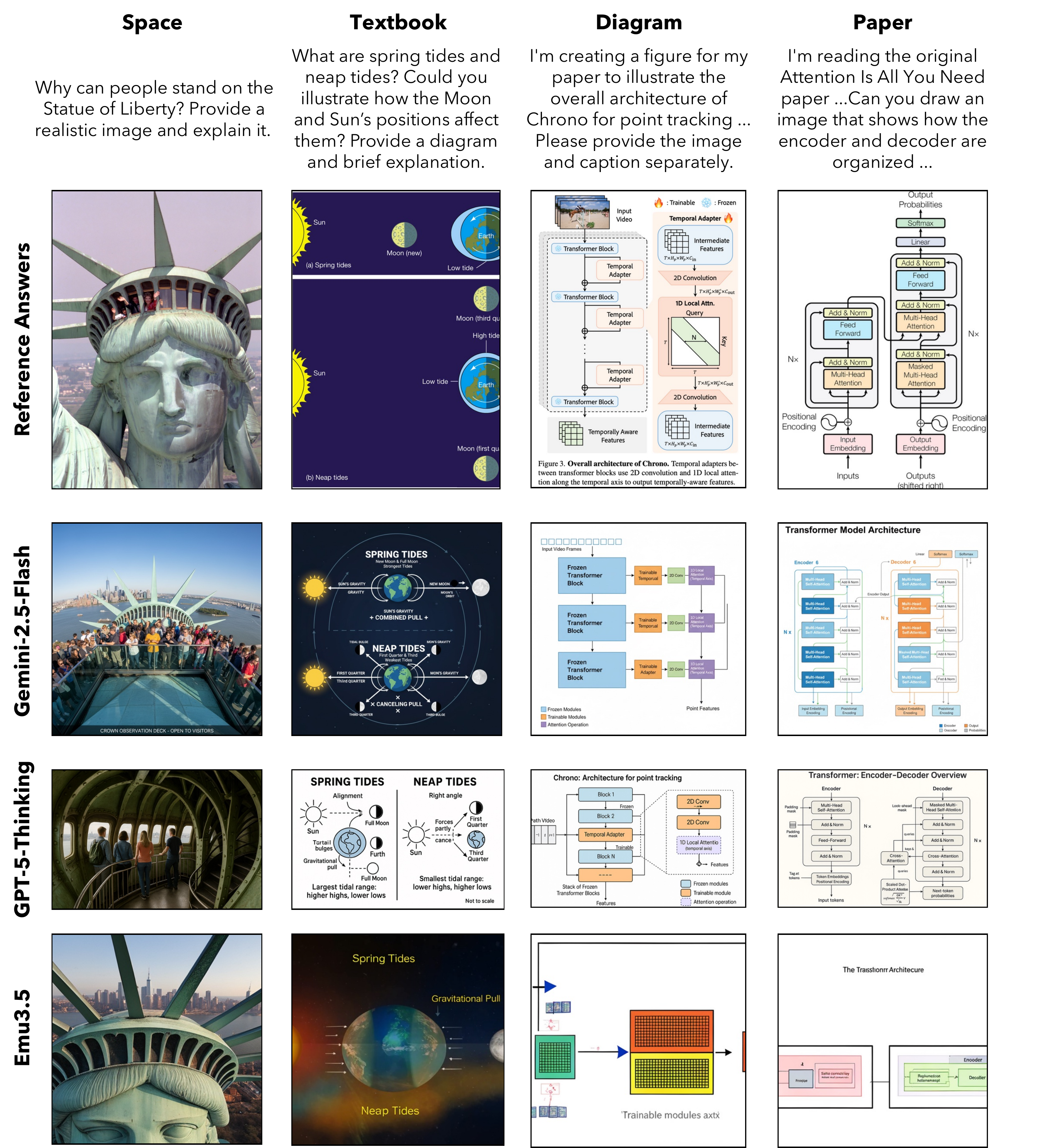}
  \caption{\paragraph{Model-generated images for closed-ended tasks in \benchmarkname{}} We visualize images generated by GPT-5-Thinking, Gemini-2.5-Flash, and Emu3.5~\citep{cui2025emu35nativemultimodalmodels}. These images fail to answer the questions accurately. For example, Gemini-2.5-Flash depicts a nonexistent external platform above the Statue of Liberty instead of the crown interior.}
  \label{fig:model_responses_closed}
\end{figure}

\begin{figure}[t]
  \centering
  \includegraphics[width=\linewidth]{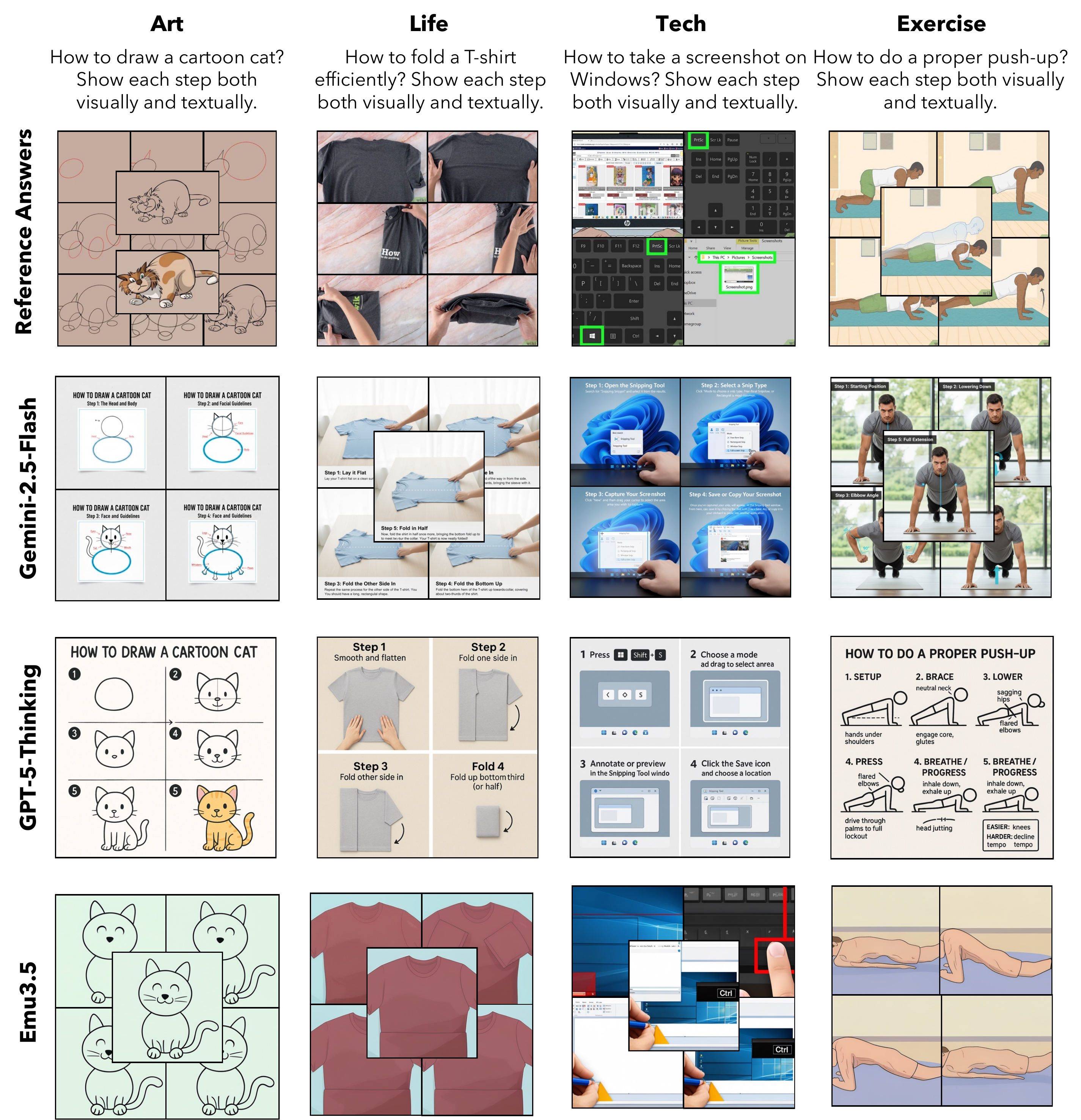}
  \caption{\paragraph{Model-generated images for open-ended tasks in \benchmarkname{}} We prompt GPT-5-Thinking, Gemini-2.5-Flash, and Emu3.5 to synthesize step-by-step visual guides for each task. The generated images often exhibit temporal inconsistencies. For instance, in the \textit{art} task, when drawing a cartoon cat, GPT-5-Thinking mislabels sub-images (\eg, two images tagged as step 5). For visualization, we stack the images generated by Gemini-2.5-Flash into a single grid.}
  \label{fig:model_responses_open}
\end{figure}
\section{\benchmarkname{}}

We present \benchmarkname{}, a benchmark designed to evaluate \textit{unified multimodal} generation. \benchmarkname{} focuses on real-world tasks where a model must reason carefully before generating natural language and images in response to user queries. \benchmarkname{} consists of 8 tasks. These tasks fall into two groups: closed-ended tasks and open-ended tasks. They differ in both task design and evaluation dimensions used for model responses.

Closed-ended tasks, including \textit{space}, \textit{textbook}, \textit{diagram}, and \textit{paper}, emphasize factual understanding and grounded explanation, and typically have a clear target answer that model outputs are expected to match. In contrast, open-ended tasks, including \textit{art}, \textit{life}, \textit{tech}, and \textit{exercise}, focus on step-by-step drawings that illustrate how to perform a task. Multiple plausible visualizations may exist for the same question in these tasks.

Figures~\hyperref[fig:model_responses_closed]{\textcolor{link}{\ref*{fig:model_responses_closed}}} and~\hyperref[fig:model_responses_open]{\textcolor{link}{\ref*{fig:model_responses_open}}} illustrate the 8 tasks in {\benchmarkname{}} alongside model-generated images. These tasks range from explanations with visual illustrations (\eg, space) to academic figure creation (\eg, paper). They also vary in format, from multi-step generation to single-step description (guide \vs diagram), and in breadth, from general scientific knowledge to specialized academic content (textbook \vs paper). 

Each sample in our benchmark includes a question prompt and a grading rubric with multiple rubric criteria to score model outputs. Our rubric criteria are drafted by Gemini-2.5-Pro~\citep{comanici2025gemini25pushingfrontier} and then refined by humans (see \secref{sec:evaluation} for more details). During evaluation, these rubrics are used by an MLLM judge to grade model responses for each question. In total, \benchmarkname{} contains 1,000 questions and 10,417 rubric criteria.

\subsection{Dataset Composition}
\label{sec:dataset}

We describe each task in {\benchmarkname{}}. All open-ended tasks, including \textit{art}, \textit{life}, \textit{tech}, and \textit{exercise}, are grouped under a broader category, \textit{guide}, as they share task design and data-collection procedure. Additional details on the image and text sources used to build each task are provided in \appendixref{sec:source}.

\paragraph{Space} This task evaluates a model's ability to depict specific architectural features. Generated images must highlight the structural elements relevant to the question rather than serve as decoration. For example, given the prompt \textit{``Why can people stand on the Statue of Liberty? Provide a realistic image and explain it.''}, a full-view image of the statue is insufficient, as it does not show the crown platform where visitors can stand. 

To construct this task, we first collect a small set of 20 seed questions about well-known landmarks from online Q\&A forums (\eg, Quora-like platforms). Since there are not many such questions on the internet or in existing datasets, we use GPT-5 to expand this set. Given our seed questions, the model is prompted to propose additional questions for different landmarks. Human annotators then review all generated questions to ensure that each one refers to real, identifiable features of a landmark.

For every verified question, annotators retrieve a representative image from public sources (\eg, Wikipedia) that depicts the relevant architectural feature and can answer the question visually. Finally, annotators write a short reference text describing how the chosen image illustrates the engineering features of that landmark.

\paragraph{Textbook} This task tests a model's ability to explain fundamental scientific phenomena (\eg, geological transformations) through instructional diagrams. Generated outputs should identify underlying mechanisms (\eg, DNA's role in genetics) or connections between concepts (\eg, how a headland erodes into caves, arches, and stacks). An example question is \textit{``I do not quite get how a headland turns into caves, arches, and stacks. Can you generate an image and explain the sequence? Please answer with both visual and textual explanations.''}

We use the textbook-style diagrams and their answer texts in the TQA dataset~\citep{kembhavi2017you} as our reference image-text pairs. These pairs cover several subjects in science, including biology, geography, and chemistry. Building on these references, we prompt GPT-5 to generate learning-oriented questions that can be answered using our reference answers. Human annotators then manually review the generated questions to ensure that each is scientifically sound, unambiguous, and can be answered with the reference content. 

\paragraph{Diagram} This task targets a common need in academic writing, where researchers design figures to illustrate complex methods in research papers. The evaluated model receives a technical description of a specific method or model architecture and is asked to synthesize a self-contained figure with a caption.

To construct this task, we manually curate figures from recent top-tier AI conference papers (\eg, ICLR, CVPR) and pair each figure with its original caption as a reference image-text pair. We intentionally avoid diagrams from well-known papers to reduce the chance that evaluated models have seen them during training. We provide GPT-5 with the image as well as the full paper and prompt it to create a figure-generation instruction. This instruction captures important architectural or methodological details so that models can answer the question without needing access to the original paper. Human annotators then review and refine the instructions to check scientific correctness and relevance to the original figure. 

\paragraph{Paper} This task assesses whether a unified model can accurately explain complex concepts from cutting-edge computer science research in an accessible way. Given a user question about a particular method, the model must first understand the technical content and then provide a clear, coherent explanation. An example question is \textit{``I'm reading the original Attention Is All You Need paper and trying to understand the overall structure of the Transformer model. Can you draw an image that shows how the encoder and decoder are organized, and explain how data flows through the layers? Please answer with both visual and textual answers.''}

We source a diverse set of figures from seminal papers (\eg, Transformer~\citep{vaswani2017attention}, ResNet~\citep{he2016deep}) as well as online teaching platforms (\eg, D2L~\citep{zhang2023dive}). Each figure serves as a reference image. For each reference image, we extract the relevant technical descriptions from the original sources and prompt GPT-5 to generate reader-oriented questions together with explanatory reference text that answers those questions based on the figure. Human annotators then review the generated questions and answers to check technical validity and faithfulness to the original content.

\paragraph{Guide} This task evaluates a model's ability to produce a coherent, step-by-step visual guide for everyday activities. It contains four tasks, \textit{art}, \textit{life}, \textit{tech}, and \textit{exercise}, covering a range of real-world skills that require multi-step demonstration. For each question, the model must generate a visual guide (in one or more images) together with text explanations that illustrate a clear progression from an initial state to the final state. An example question from these tasks is \textit{``How to draw a cartoon cat? Show each step both visually and textually.''}

Questions and reference image-text answers are sourced from high-quality tutorial materials (\eg, WikiHow) and step-by-step demonstrational videos (\eg, YouTube). Because these tasks require visually illustrating specific skills through multi-step drawings, each reference answer consists of a sequence of images that progressively depict the drawing process and stepwise textual explanations. For the questions, we do not specify the number of steps and allow models to flexibly produce a sequence of images for each task.

\subsection{Rubric Generation and Evaluation}
\label{sec:evaluation}

How to effectively evaluate unified multimodal generation remains an open problem. \citet{zhou2025opening} use average win rates from pairwise comparisons of model outputs containing images and text, but this approach requires a large number of comparisons to obtain accurate model scores. These scores can change if a different set of models is evaluated. Moreover, most current evaluations~\citep{xia2024mmie,zhou2025opening} are \textit{data-independent}: a single generic prompt is applied to grade all samples. As a result, this approach overlooks sample-specific differences and can lead to inaccurate results. 

\begin{figure}[t]
    \centering 
    \includegraphics[width=\linewidth]{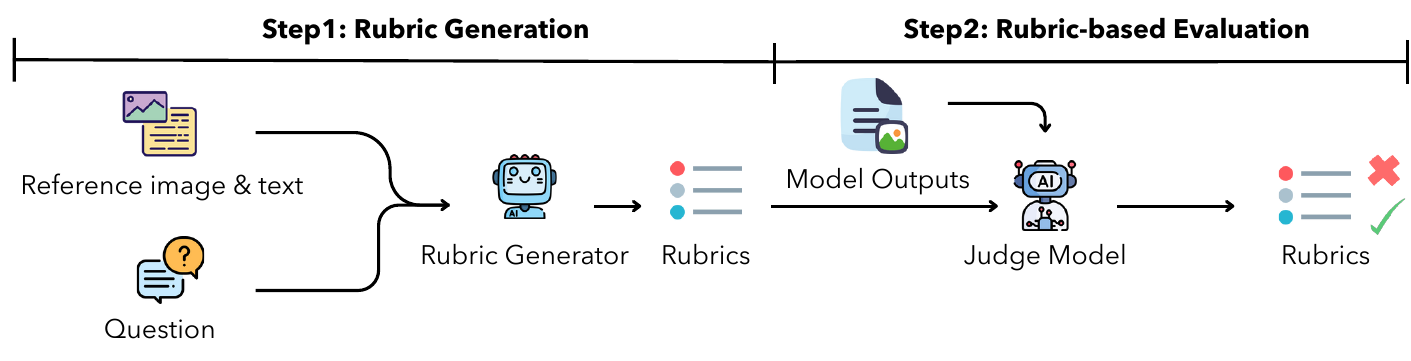}
    \caption{\paragraph{Rubric drafting and response evaluation procedure in \benchmarkname{}} We propose using \textit{data-dependent} rubrics to evaluate outputs from unified models. For each question, a model drafts an itemized rubric based on the question and the reference image-text pair. A judge model then scores the generated response against each rubric criterion.}
    \label{fig:evaluation}
\end{figure}
\paragraph{Rubric generation} Inspired by HealthBench~\citep{arora2025healthbenchevaluatinglargelanguage}, we adopt a \textit{data-dependent} approach to evaluate unified multimodal generation. \figref{fig:evaluation} (\textit{top}) illustrates our framework for generating rubrics and evaluating model outputs with an MLLM. For each sample, we provide the question along with its reference image-text answer to a rubric generator (\eg, Gemini-2.5-Pro) to create a rubric with multiple fine-grained criteria that evaluate model outputs along different dimensions. For each question, about half of the rubric criteria are used to evaluate the generated image(s), while the others are used to evaluate the generated text.

\begin{figure}[t]
    \centering 
    \includegraphics[width=\linewidth]{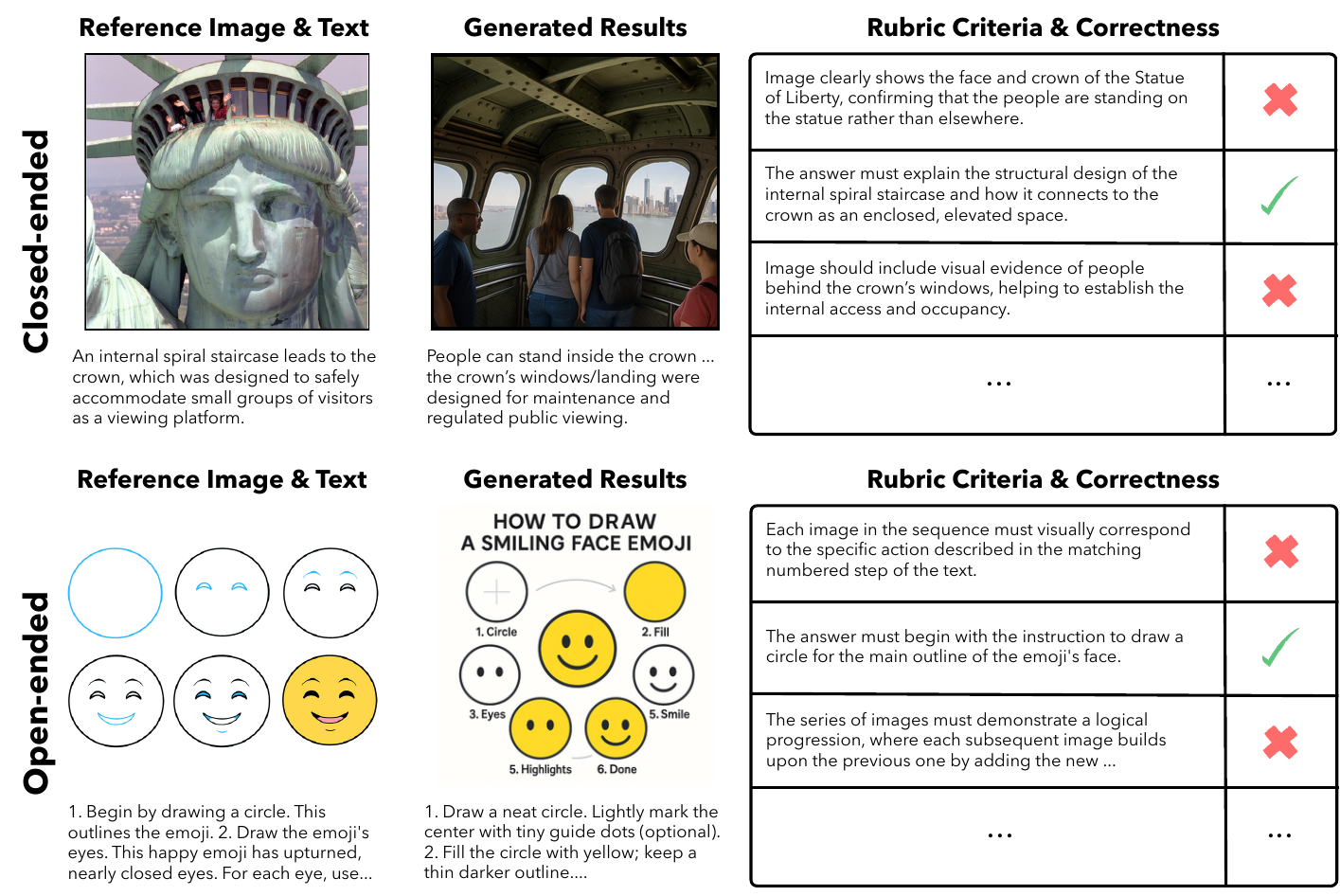}
    \caption{\paragraph{Rubric examples for closed-ended and open-ended tasks in \benchmarkname{}} Rubrics for closed-ended tasks check whether generated outputs contain specific details important for answering the questions, whereas rubrics for open-ended tasks evaluate higher-level generation qualities (\eg, image-text alignment).}
    \label{fig:rubric_grading}
\end{figure}

Note that we prompt the rubric generator differently when drafting rubrics for closed-ended \vs open-ended tasks. \figref{fig:rubric_grading} illustrates this difference. For closed-ended tasks, rubrics check whether the generated content contains the key information present in the reference answers. In contrast, for open-ended tasks, rubrics evaluate higher-level qualities of the output (\eg, temporal consistency). Rubrics for open-ended tasks include criteria such as \emph{``The answer must describe a step-by-step process for drawing a cat''} and \emph{``The sequence of images must illustrate the complete drawing process, from the initial basic shapes to the final version''}. 

\paragraph{Human review} We conduct two rounds of human review to ensure the quality and reliability of all generated rubrics. Our goal is to design rubrics that correctly reward responses similar to the reference answers while penalizing erroneous ones. For each benchmark sample, a primary annotator first verifies the question, reference answer, and rubrics for correctness and alignment with the task. Subsequently, other co-authors independently review the annotations. Only rubric items unanimously judged by all reviewers to be unambiguous and well aligned with the task design are retained. Overall, human annotators supervise benchmark construction at a system level, from the inputs (questions) to the outputs (reference answers) and the grading criteria (rubrics). 

Human annotators refine the model-generated rubric items in several ways. First, repeated or similar rubric criteria are consolidated. For example, we merge \textit{``the steps must be sequential''} and \textit{``the steps should follow a logical order''} into a single, more precise requirement. Second, we add important but missing rubric criteria. For example, in questions involving rendered text, we introduce additional rubric criteria to evaluate overall text generation quality, such as \textit{``all visible text in the generated image(s) must be spelled correctly and rendered naturally, with no misspellings, garbled characters, distortions, or nonsensical text''}. 

\paragraph{Rubric-based evaluation} We employ an MLLM (\eg, Gemini-2.5-Pro) as a judge to evaluate images and text generated by unified models. For each sample, the judge checks the model response against each rubric criterion, and the final score is computed as the fraction of satisfied rubric criteria over the total number of rubric criteria. This provides an automated, reproducible evaluation method in place of human judgment. We find that using a frontier model yields results well aligned with human judgment, and that some strong judge models produce similar scores. We will discuss this further in \secref{sec:analysis}.

In \tabref{tab:task_results_all} (\emph{\textcolor{gray}{gray}}), we report the scores of the reference answers graded with our rubrics, using Gemini-2.5-Pro as the judge model. Overall, the reference image-text answers achieve an average rubric score of 92.2 across all tasks, indicating that our rubrics capture most of the important features of the reference answers. Nevertheless, we observe that reference answers for open-ended tasks receive slightly lower scores than those for closed-ended tasks. This is likely because reference answers for open-ended tasks contain multiple images, making them harder for the judge model to evaluate accurately, whereas closed-ended tasks involve only a single image. We expect that more capable MLLMs will further improve the effectiveness of our rubric-based evaluation.

\section{Experiments}

\subsection{Settings}
\label{sec:setup}

\paragraph{Models} We evaluate recent unified models on all 8 tasks in our benchmark. For open-source models, we consider Janus-Pro~\citep{chen2025janus}, Show-o2~\citep{xie2025show}, MMaDA~\citep{yang2025mmada}, BAGEL~\citep{deng2025emerging}, and Emu3.5~\citep{cui2025emu35nativemultimodalmodels}. For proprietary frontier models, we evaluate Gemini-2.0-Flash~\citep{gemini2imagegen}, Gemini-2.5-Flash (\ie, Nano Banana)~\citep{nanobanana}, GPT-5-Instant~\citep{gpt5}, and GPT-5-Thinking~\citep{gpt5}. We access both GPT-5 models through the official chat interface.

\paragraph{Evaluation setup} Some models (\eg, Janus-Pro, Show-o2, MMaDA, and BAGEL) can generate only images or text by design, but not both in a single inference pass. To obtain both outputs, we feed the same question prompt to the model twice, once to generate the image and once to generate the text. For models that natively support joint image-text generation (\eg, GPT-5, Gemini), we directly collect their multimodal responses. We use Gemini-2.5-Pro~\citep{comanici2025gemini25pushingfrontier} to grade the generated outputs based on fine-grained rubrics (\secref{sec:evaluation}). \appendixref{sec:prompt} provides the full prompt to evaluate model responses.

\begin{table}[t]

\centering
\tablestyle{3pt}{1.2}
\begin{tabular}{lx{38.5}x{38.5}x{38.5}x{38.5}x{38.5}x{38.5}x{38.5}x{38.5}x{38.5}}
\shline
& \textbf{Space}
& \textbf{Textbook}
& \textbf{Diagram}
& \textbf{Paper}
& \textbf{Art}
& \textbf{Life}
& \textbf{Tech}
& \textbf{Exercise}
& \textbf{Avg} \\
\Xhline{0.7pt}
\textcolor{gray}{Reference} & \textcolor{gray}{96.2} & \textcolor{gray}{94.4} & \textcolor{gray}{93.1} & \textcolor{gray}{96.2} & \textcolor{gray}{90.6} & \textcolor{gray}{87.7} & \textcolor{gray}{90.6} & \textcolor{gray}{89.2} & \textcolor{gray}{92.2} \\
\Xhline{0.7pt}
\multicolumn{10}{c}{\textit{Open-source Models}} \\
Janus-Pro & 21.0 & 31.0 & 37.4 & 15.2 & 26.4 & 23.0 & 17.6 & 11.5 & 22.9 \\
Show-o2 & 25.4 & 33.1 & 33.2 & 17.4 & 25.6 & 15.6 & 17.4 & 13.1 & 22.6 \\
MMaDA & 10.8 & 20.0 & 14.2 & 13.3 & 15.7 & 15.8 & 12.4 & 12.6 & 14.4 \\ 
BAGEL & 29.8 & 42.5 & 37.2 & 20.0 & 39.0 & 33.6 & 24.8 & 21.4 & 31.0\\
Emu3.5 &\textbf{59.1} & \textbf{57.4} & \textbf{41.1} & \textbf{31.6} & \textbf{59.3} & \textbf{62.0} & \textbf{37.0} & \textbf{45.4 }& \textbf{49.1}   \\
\Xhline{0.7pt}
\multicolumn{10}{c}{\textit{Proprietary Frontier Models}} \\
Gemini-2.0-Flash & 65.2 & 55.2 & 47.6 & 45.8 & \textbf{70.4} & 58.0 & 50.2 & 48.0 & 55.1 \\
Gemini-2.5-Flash  & 78.0 & 74.0 & 66.4 & \textbf{71.6} & 66.6 & 63.0 & \textbf{58.2} & 50.0 & 66.0 \\
GPT-5-Instant   & 77.3 & 77.9 & 62.3 & 55.1 & 71.2 & \textbf{69.7} & 50.7 & 57.6 & 65.2 \\
GPT-5-Thinking  & \textbf{84.0} & \textbf{78.0} & \textbf{67.8} & 51.9 & 67.8 & 63.8 & 57.0 & \textbf{61.4} & \textbf{66.4} \\
\shline
\end{tabular}

\caption{\paragraph{\benchmarkname{} leaderboard} We evaluate open-source and proprietary frontier models on 8 tasks in \benchmarkname{}. \textbf{Bold} indicates the highest performance for each column within each group (\eg, open-source \vs proprietary frontier).}
\label{tab:task_results_all}
\end{table}

\subsection{Results}
\label{sec:results}

\begin{figure}[t]
    \centering
    \begin{minipage}[t]{.49\linewidth}
    \centering
        \includegraphics[width=\linewidth]{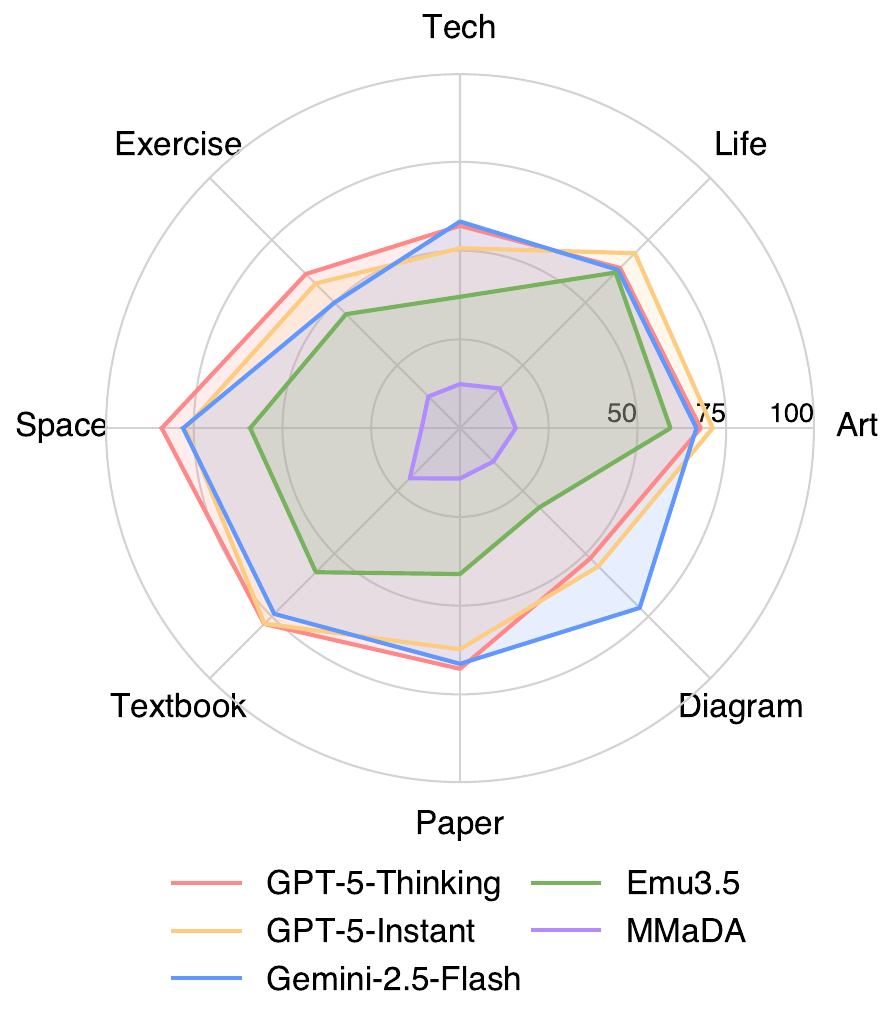}
        \caption{\paragraph{Model performance on \benchmarkname{}} Each axis corresponds to a task, and each colored polygon is a different model. Proprietary frontier models (\eg, GPT-5-Thinking) consistently outperform open-source models (\eg, Emu3.5).}~\label{fig:model_radar}
    \end{minipage}%
    \hfill
    \begin{minipage}[t]{.49\linewidth}
    \centering
        \vspace{-29.0em}
        \includegraphics[width=\linewidth]{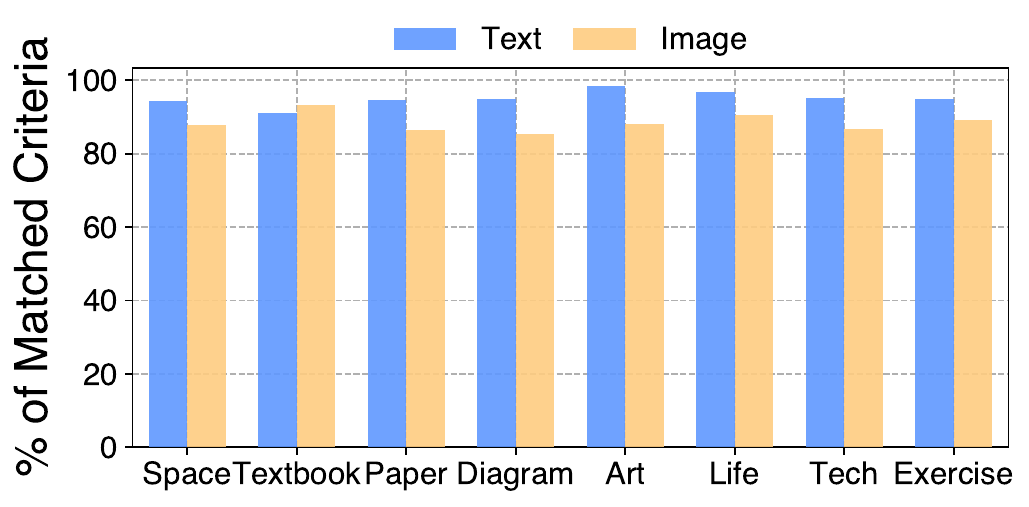}
        \vspace{-1.5em}
        \caption{\paragraph{Percentage of matched rubric criteria between human evaluators and an MLLM judge} We find high agreement between an MLLM and humans.}~\label{fig:matching_criteria}
        \vspace{.15em}
        \includegraphics[width=\linewidth]{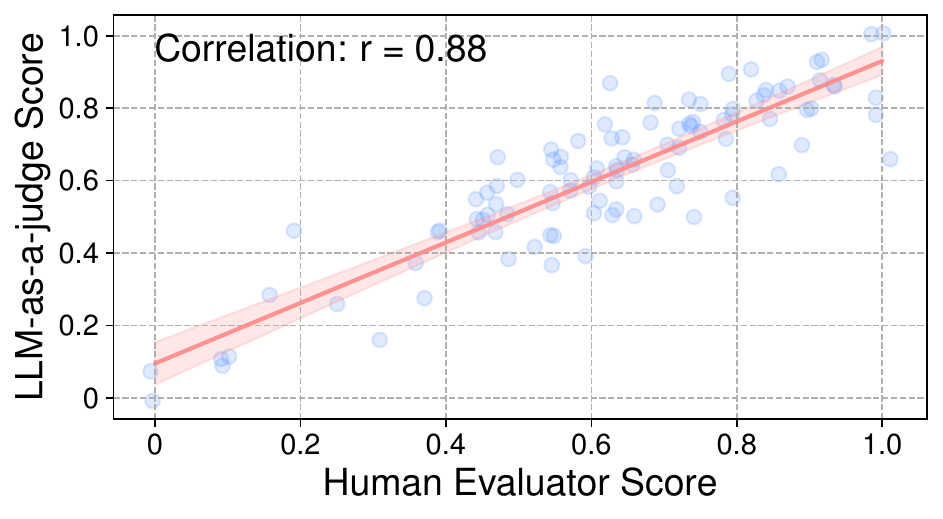}
        \vspace{-1.5em}
        \caption{\paragraph{LLM-as-a-judge \vs human evaluator} Each point represents a score for a question in {\benchmarkname{}}. The Pearson correlation (r = 0.88) demonstrates that the LLM judge closely aligns with human evaluation.}~\label{fig:correlation}
    \end{minipage}
\end{figure}

\tabref{tab:task_results_all} reports the performance of various models on \benchmarkname{}. Overall, frontier models consistently outperform open-source ones across all tasks: GPT-5-Thinking achieves the highest average score of 66.4, while the best open-source model obtains only 49.1. To better understand performance differences, \figref{fig:model_radar} presents a radar chart comparing models across tasks. The gap between proprietary and open-source models is large: the strongest frontier model (\ie, GPT-5-Thinking) outperforms the best open-source model (\ie, Emu3.5) by over 17 points on average. The individual image and text scores for each task are provided in \appendixref{sec:individual_scores}.

We also observe that the tasks requiring multi-step planning (\eg, art, life) yield substantially lower scores than knowledge-based tasks (\eg, textbook, diagram). \figref{fig:model_responses_open} further illustrates this pattern. For example, in the \textit{art} task, GPT-5-Thinking incorrectly labels the final two images as \textit{step 5}. Similar mistakes also occur in both the \textit{life} and \textit{exercise} tasks. Likewise, Gemini-2.5-Flash changes the shirt's orientation from step 1 to step 2 and then changes it back in step 3 in the \textit{life} task. More interestingly, current \textit{reasoning} models (\eg, GPT-5-Thinking) achieve much better performance than non-reasoning ones (\eg, GPT-5-Instant). We further study the benefits of \textit{reasoning} in multimodal generation in the next section.

\subsection{Analysis}
\label{sec:analysis}
\begin{wraptable}{r}{0.35\linewidth}
\vspace{-1em}
\centering
\tablestyle{3pt}{1.2}
\begin{tabular}{l x{38.5} x{38.5}}

\shline
& \textbf{Image} & \textbf{Text} \\
\Xhline{0.7pt}

\textcolor{gray}{Reference}
& \textcolor{gray}{87.8}
& \textcolor{gray}{96.8} \\
\Xhline{0.7pt}
\multicolumn{3}{c}{\textit{Open-source Models}} \\
Janus-Pro  &  6.5 & 39.3 \\
Show-o2   & 8.3 & 36.9 \\
MMaDA    & 15.9 & 12.8 \\
BAGEL     & 13.6 & 48.5 \\
Emu3.5    & \textbf{33.6} & \textbf{64.6} \\

\Xhline{0.7pt}
\multicolumn{3}{c}{\textit{Proprietary Frontier Models}} \\
Gemini-2.0-Flash  & 36.9 & 73.2  \\
Gemini-2.5-Flash  & \textbf{56.4} & 75.5 \\
GPT-5-Instant     & 52.8 & 77.7 \\
GPT-5-Thinking    & 49.1 & \textbf{83.8 }\\

\Xhline{0.7pt}
\multicolumn{3}{c}{\textit{Text-only Models}} \\
Qwen3-32B & -- & 81.2 \\
Gemma3-27B & -- & 83.1 \\
\shline
\end{tabular}
\caption{\paragraph{Image and text scores on UEval, averaged over all 8 tasks} We find a performance gap between image and text generation. We also report the text scores of text-only models for comparison.}
\label{tab:image_text_score}
\vspace{-2em}
\end{wraptable}

\paragraph{Performance on text and image generation}
As described in \secref{sec:evaluation}, each question contains multiple rubric criteria, with about half evaluating generated images and the others evaluating generated text. \tabref{tab:image_text_score} reports separate image and text scores for all evaluated models (from \tabref{tab:task_results_all}), averaged across all questions to assess text and image generation performance separately. We also report text scores for text-only models. Overall, current models show stronger performance in text generation than in image generation. Moreover, for open-source unified models, their text generation ability still lags behind that of text-only models, highlighting that achieving high-quality multimodal generation is still a significant challenge and that substantial room for improvement remains.

\paragraph{Human evaluation} To assess the reliability of using an MLLM as a judge, we randomly sample 10\% of GPT-5-Thinking outputs from each task and ask human annotators to determine how many rubric criteria each model response satisfies. In \figref{fig:matching_criteria}, we report the percentage of rubric criteria on which human evaluators and the LLM-as-a-judge agree. We observe approximately 90\% of rubric criteria are matched between human evaluators and the LLM-as-a-judge across tasks. LLM-as-a-judge grading aligns very closely with human judgment across different tasks. Moreover, in \figref{fig:correlation}, we plot LLM-as-a-judge scores against human evaluator scores and observe a high Pearson correlation (r = 0.88). These results indicate that our automated evaluation framework is robust aligned with human judgment.

\paragraph{Different judge models} Our default judge model in \tabref{tab:task_results_all} is Gemini-2.5-Pro~\citep{comanici2025gemini25pushingfrontier}. To understand how different judge models affect scores, we grade model responses generated by GPT-5-Thinking using other proprietary frontier and open-source models, including GPT-5-Thinking~\citep{gpt5}, Seed1.6-Vision~\citep{doubao}, Qwen3-VL-235B-Thinking/Instruct~\citep{qwen3}, and GLM-4.1V-Thinking~\citep{vteam2025glm45vglm41vthinkingversatilemultimodal}. 

\begin{table*}[h]
\centering
\tablestyle{4pt}{1.2}
\begin{tabular}{l
x{38.5}x{38.5}x{38.5}x{38.5}
x{38.5}x{38.5}x{38.5}x{38.5}}
\shline
& \textbf{Space}
& \textbf{Textbook}
& \textbf{Diagram}
& \textbf{Paper}
& \textbf{Art}
& \textbf{Life}
& \textbf{Tech}
& \textbf{Exercise} \\
\Xhline{0.7pt}

Gemini-2.5-Pro & 84.0 & 78.0 & 67.8 & 51.9 & 67.8 & 63.8 & 57.0 & 61.4 \\
GPT-5-Thinking & 80.0 & 73.0 & 55.8 & 46.3 & 56.4 & 50.1 & 45.4 & 51.7 \\
Seed1.6-Vision & 85.5 & 81.0 & 68.2 & 53.8 & 75.2 & 70.8 & 67.8 & 70.9 \\
Qwen3-VL-235B-Thinking & 81.8 & 78.4 & 63.4 & 49.3 & 56.8 & 56.8 & 53.6 & 59.6 \\
Qwen3-VL-235B-Instruct & 82.0 & 85.2 & 72.3 & 53.6 & 73.7 & 61.4 & 57.7 & 63.0 \\
GLM-4.1V-Thinking & 84.3 & 83.6 & 68.2 & 49.8 & 79.4 & 74.5 & 74.3 & 70.7\\
\shline
\end{tabular}
\caption{\paragraph{Scores of GPT-5-Thinking responses evaluated by different judge models} Gemini-2.5-Pro, GPT-5-Thinking, and Qwen3-VL-235B-Thinking produce consistent scores across tasks, whereas other models yield very different scores.}
\label{tab:judge_value}
\end{table*}

\tabref{tab:judge_value} reports the per-task scores on GPT-5-Thinking responses graded with different judge models. We observe that GPT-5-Thinking, Gemini-2.5-Pro, and Qwen3-VL-235B-Thinking produce similar scores across all tasks, whereas other models (\eg, Seed1.6-Vision, GLM-4.1V-Thinking) yield very different ones, especially in open-ended tasks (\eg, \emph{art}, \emph{life}). Therefore, we recommend using GPT-5-Thinking, Gemini-2.5-Pro, or Qwen3-VL-235B-Thinking as the judge model for grading model-generated responses in {\benchmarkname{}}.
\begin{figure}[!b]
    \centering
    \includegraphics[width=\linewidth]{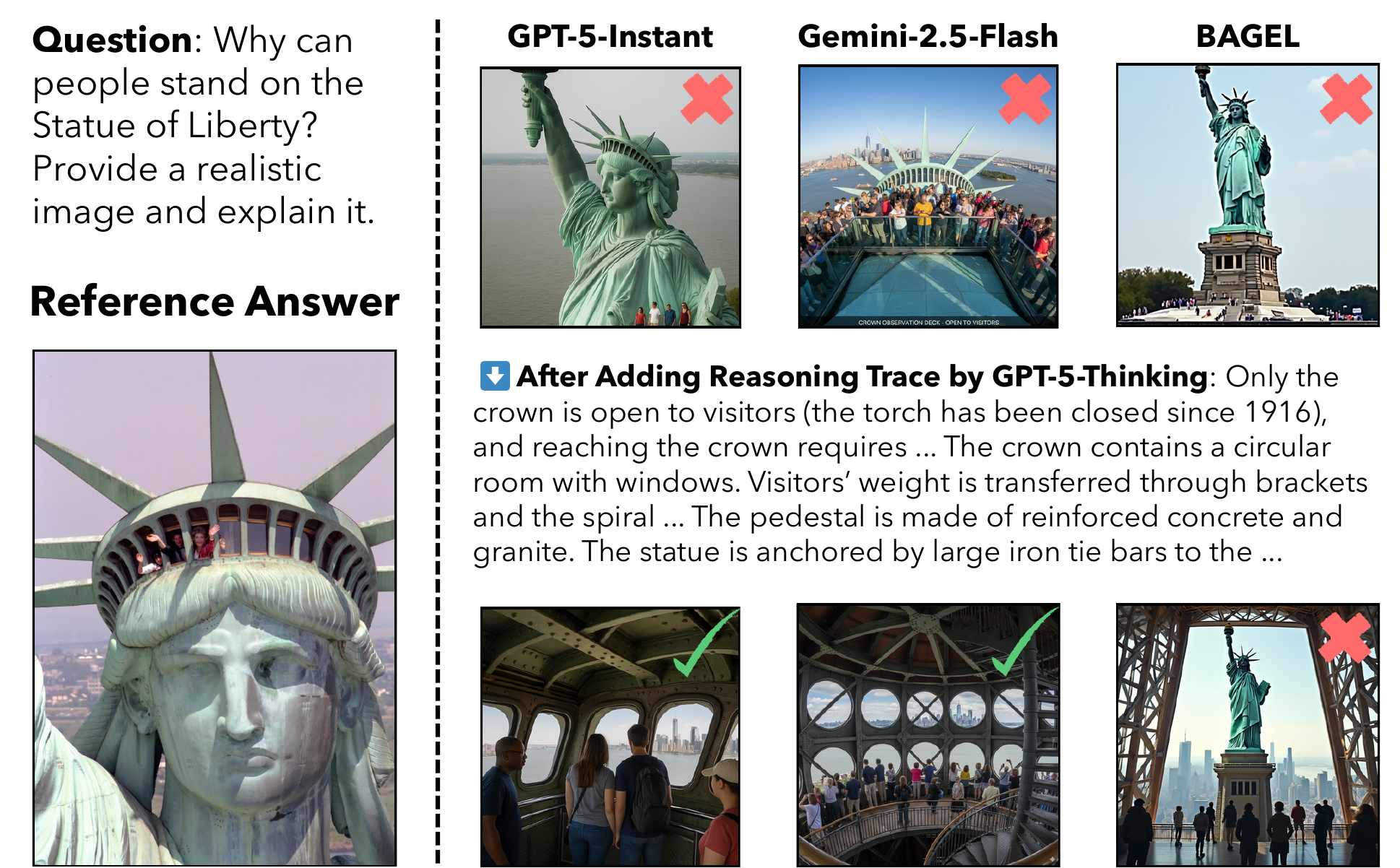}
    \caption{\paragraph{Reasoning can improve the quality of multimodal generation} We extract the \textit{reasoning trace} from GPT-5-Thinking and append it to the original question. The resulting prompt is then provided to \textit{non-reasoning} models (\eg, GPT-5-Instant, Gemini-2.5-Flash, and BAGEL) to generate responses. This can steer the generated images toward that of the reasoning model and lead to more accurate responses, with BAGEL as an exception.}
    \label{fig:thinking_analysis}
\end{figure}
\paragraph{The effectiveness of reasoning traces} To understand why reasoning models (\eg, GPT-5-Thinking) yield better results in multimodal generation, we record a reasoning trace and append it to the original question prompt. We then provide non-reasoning models (\eg, GPT-5-Instant) with this modified prompt. \figref{fig:thinking_analysis} visualizes images generated by some non-reasoning models. Surprisingly, incorporating the reasoning trace enables GPT-5-Instant and Gemini-2.5-Flash to generate a more accurate image of the interior of the Statue of Liberty’s crown. This suggests that multimodal generation in unified models benefits from Chain-of-Thought~\citep{wei2023chainofthoughtpromptingelicitsreasoning} reasoning generated by other models. However, this does not apply to all unified models: weaker models (\eg, BAGEL) do not benefit from this added reasoning. A sufficiently strong multimodal generation capability is necessary to effectively leverage such additional reasoning signals.
\section{Related Work}

\paragraph{Multimodal large language models and unified models} 
Multimodal Large Language Models (MLLMs) have progressed greatly in recent years. These models~\citep{alayrac2022flamingo, dai2023instructblip, liu2023visual, li2023blip, zhuminigpt} typically integrate a visual encoder~\citep{radford2021learningtransferablevisualmodels, dosovitskiy2020image, he2022masked} with a pre-trained Large Language Model~\citep{touvron2023llama2openfoundation,peng2023instruction}, achieving strong performance on image captioning~\citep{chen2015microsoft, plummer2015flickr30k} and visual question answering~\citep{goyal2017making, mathew2021docvqadatasetvqadocument}. To further scale their capabilities, these models require millions of instruction-tuning data~\citep{tong2024cambrian, deitke2024molmopixmoopenweights}.

A parallel line of research seeks to unify multimodal understanding and generation within a single model. Some methods adopt diffusion-based approaches~\citep{dong2024dreamllmsynergisticmultimodalcomprehension, yang2025mmada,li2025dual, shi2025muddit, wang2025fudoki}, whereas others train models with a purely autoregressive objective~\citep{team2024chameleon, wang2024emu3, wu2025janus, wu2025vilauunifiedfoundationmodel, qu2025tokenflow}. There are also hybrid methods that combine both approaches~\citep{deng2025emerging, zhou2024transfusionpredicttokendiffuse, xie2025show, chen2025blip3ofamilyfullyopen}. We refer readers to~\citet{zhang2025unifiedmultimodalunderstandinggeneration} for a comprehensive survey of MLLMs and unified models.

\paragraph{Multimodal benchmarks} A range of benchmarks has been proposed to evaluate multimodal inputs. Initial work~\citep{goyal2017making,marino2019okvqavisualquestionanswering,masry2022chartqa} evaluates image understanding for specific image types, and later efforts benchmark broader image coverage~\citep{liu2024mmbench,yue2024mmmumassivemultidisciplinemultimodal}. There are also studies evaluating text-to-image generation quality~\citep{saharia2022photorealistic,huang2023t2i,ghosh2023geneval,lin2024evaluating}. Some benchmarks (\eg, VDC~\citep{chai2024auroracap}) begin to use data-dependent rubrics for better evaluation. More recent works unify understanding and generation benchmarks as interleaved text-and-image generation~\citep{an2024openleaf,liu2024holistic,chen2024interleaved} or unified multimodal generation~\citep{li2025unieval, zou2025uni, shi2025realunifyunifiedmodelstruly}. In contrast to them, our evaluation is very simple: an MLLM judge is used to grade model responses based on rubrics. This avoids per-sample human scoring~\citep{zhou2025opening} or training a scoring model~\citep{xia2024mmie} for evaluation.
\section{Conclusion}

We introduce \benchmarkname{}, a benchmark to evaluate unified multimodal generation beyond standard tasks (\eg, visual question answering, text-to-image generation). Our benchmark contains 1,000 samples across 8 real-world tasks and provides 10,417 fine-grained rubric criteria for rigorous, automated grading of model responses. Our results demonstrate that \benchmarkname{} is challenging for both proprietary frontier and open-source unified models. We also observe that reasoning can improve multimodal generation quality. We hope this work will stimulate further research on developing stronger models and better benchmarks for multimodal generation.

\section{Acknowledgments}

We gratefully acknowledge the use of the Neuronic GPU computing cluster maintained by the Department of Computer Science at Princeton University. This work was performed using Princeton Research Computing resources, a consortium led by the Princeton Institute for Computational Science and Engineering (PICSciE) and Research Computing at Princeton University. 
\clearpage

\bibliography{main}
\bibliographystyle{main}

\clearpage
\newpage
\appendix
\section*{\Large{Appendix}}

\section{Individual Image and Text Scores}
\label{sec:individual_scores}

As detailed in \secref{sec:evaluation}, we construct two rubrics for every question in {\benchmarkname{}}, one evaluating images and one evaluating text. Table~\hyperref[tab:task_results_image_text]{\textcolor{link}{\ref*{tab:task_results_image_text}}} reports the corresponding per-modality scores for the results in \tabref{tab:task_results_all}. Across all models, we observe that text scores are much higher than image scores. This is especially pronounced for open-source models, which generate reasonably good text but perform extremely poorly on image generation, often receiving <10 image scores. In contrast, frontier models show a much narrower gap between their image and text performance, indicating more balanced multimodal generation capabilities.

\begin{table}[h]
\centering
\tablestyle{3.3pt}{1.15}
\begin{tabular}{lx{33}x{33}x{33}x{33}x{33}x{33}x{33}x{33}x{33}x{33}}
\shline
& \multicolumn{2}{c}{\textbf{Space}}
& \multicolumn{2}{c}{\textbf{Textbook}} 
& \multicolumn{2}{c}{\textbf{Diagram}} 
& \multicolumn{2}{c}{\textbf{Paper}} 
& \multicolumn{2}{c}{\textbf{Avg}}\\
& Image & Text & Image & Text & Image & Text & Image & Text & Image & Text \\
\Xhline{0.7pt}
\textcolor{gray}{Reference} & \textcolor{gray}{94.7} & \textcolor{gray}{97.8} & \textcolor{gray}{92.5} & \textcolor{gray}{96.2} & \textcolor{gray}{92.2} & \textcolor{gray}{93.9} & \textcolor{gray}{94.7} & \textcolor{gray}{97.8} & \textcolor{gray}{93.5} & \textcolor{gray}{96.4} \\
\Xhline{0.7pt}
&\multicolumn{9}{c}{\textit{Open-source Models}} \\
Janus-Pro & 13.5 & 28.5 & 10.8 & 51.3 & 0.8 & 73.9 & 3.3 & 27.2 & 7.1 & 45.2  \\ 
Show-o2 & 22.7 & 28.1 & 10.9 & 55.2 & 9.3 & 57.2 & 4.6 & 30.2 & 11.9 & 42.7 \\
MMaDA & 5.0 & 16.7 & 17.7 & 22.4 & 9.2 & 19.2 & \textbf{15.6} & 11.0 & 11.9 & 17.3  \\
BAGEL & 27.6 & 31.9 & 14.5 & 70.6 & 2.8 & 71.7 & 4.8 & 35.2 & 12.4 & 52.3 \\
Emu3.5 & \textbf{64.6} & \textbf{53.6} & \textbf{32.5} & \textbf{82.3} & \textbf{13.3} & \textbf{68.9} & 12.3 & \textbf{50.8} & \textbf{30.7} & \textbf{63.9} \\
\Xhline{0.7pt}
&\multicolumn{9}{c}{\textit{Proprietary Frontier Models}} \\
Gemini-2.0-Flash & 59.9 & 70.5 & 29.9 & 80.5 & 21.9 & 73.3 & 17.3 & 74.4 & 32.2 & 74.7\\
Gemini-2.5-Flash & \textbf{82.6} & 73.5 & 61.5 & 86.6 & \textbf{55.7}& 77.2 & \textbf{59.7} & \textbf{83.4} & \textbf{64.9} & \textbf{80.2 }\\
GPT-5-Instant    & 74.9 & 79.7 & \textbf{67.4 }& 88.5 & 44.5 & 80.1 & 27.4 & 82.8 & 53.6 & 82.8 \\
GPT-5-Thinking   & 82.3 & \textbf{85.7} & 65.9 & \textbf{90.0} & 51.2 & \textbf{84.3} & 43.3 & 60.5 & 60.7 & 80.1 \\
\shline
\end{tabular}
\caption*{\centering(a) closed-ended tasks}
\begin{tabular}{lx{33}x{33}x{33}x{33}x{33}x{33}x{33}x{33}x{33}x{33}}
\shline
& \multicolumn{2}{c}{\textbf{Art}}
& \multicolumn{2}{c}{\textbf{Life}} 
& \multicolumn{2}{c}{\textbf{Tech}} 
& \multicolumn{2}{c}{\textbf{Exercise}} 
& \multicolumn{2}{c}{\textbf{Avg}}\\
& Image & Text & Image & Text & Image & Text & Image & Text & Image & Text \\
\Xhline{0.7pt}
\textcolor{gray}{Reference} & \textcolor{gray}{82.5} & \textcolor{gray}{98.8} & \textcolor{gray}{78.3} & \textcolor{gray}{97.1} & \textcolor{gray}{87.0} & \textcolor{gray}{94.2} & \textcolor{gray}{80.1} & \textcolor{gray}{98.2} & \textcolor{gray}{82.0} & \textcolor{gray}{97.1} \\
\Xhline{0.7pt}
&\multicolumn{9}{c}{\textit{Open-source Models}} \\
Janus-Pro & 8.9 & 44.0 & 5.9 & 40.0 & 4.4 & 30.8 & 4.1 & 18.9 & 5.8 & 33.4  \\ 
Show-o2 & 5.8 & 45.4 & 4.2 & 26.9 & 4.6 & 30.2 & 4.0 & 22.2 & 4.6 & 31.2  \\
MMaDA & 18.9 & 12.5 & 21.5 & 10.1 & \textbf{17.3} & 7.5 & 22.1 & 3.0 & 20.0 & 8.3 \\
BAGEL& 19.8 & 58.2 & 19.1 & 48.1 & 5.0 & 44.6 & 15.2 & 27.7 & 14.8 & 44.6 \\
Emu3.5 & \textbf{39.0} & \textbf{79.6} & \textbf{53.6} & \textbf{70.4} & 16.2 & \textbf{57.8} & \textbf{37.6} & \textbf{53.3} & \textbf{36.6} & \textbf{65.3} \\
\Xhline{0.7pt}
&\multicolumn{9}{c}{\textit{Proprietary Frontier Models}} \\
Gemini-2.0-Flash & 55.5 & 85.2 & 44.9 & 71.2 & 30.1 & 70.4 & 35.7 & 60.4 & 41.6 & 71.8  \\
Gemini-2.5-Flash & 51.2 & 81.9 & 55.2 & 70.8 & \textbf{48.9} & 67.4 & 36.6 & 63.4 & 48.0 & 70.9 \\
GPT-5-Instant   & \textbf{58.9} & 83.4 & \textbf{60.7} & 78.8 & 34.3 & 67.1 & \textbf{54.2} & 61.0 & \textbf{52.0} & 72.6 \\
GPT-5-Thinking  & 40.8 & \textbf{94.7} & 41.8 & \textbf{85.8 }& 26.2 & \textbf{87.9} & 41.2 & \textbf{81.5 }& 37.5 & \textbf{87.5}\\
\shline
\end{tabular}
\caption*{\centering(b) open-ended tasks}
\caption{\paragraph{Image and text scores on 8 tasks in \benchmarkname{}} Each task is evaluated separately for image and text generation. The results show that text generation consistently outperforms image generation.}
\label{tab:task_results_image_text}
\end{table}

\clearpage
\newpage
\section{Data sources}
\label{sec:source}
\tabref{tab:detailed_task_info_split} lists the data sources of the questions, reference images, and reference texts for each task in {\benchmarkname{}}.

\begin{table}[h]
\centering
\tablestyle{3pt}{1.15}
\begin{tabular}{ly{135}y{135}y{135}}
\shline
\textbf{Task} & \textbf{Question Source} & \textbf{Reference Image Source} & \textbf{Reference Text Source} \\
\Xhline{0.7pt}
\multirow{2}{*}{Space} &
Quora, GPT-5 generated &
Wikipedia, TripAdvisor, Google search &
Wikipedia, TripAdvisor, Quora \\
\Xhline{0.7pt}
\multirow{2}{*}{Textbook} &
GPT-5 generated &
TQA dataset &
TQA dataset \\
\\
\Xhline{0.7pt}
\multirow{2}{*}{Diagram} &
GPT-5 generated &
arXiv, ICLR, CVPR, NeurIPS, ICCV &
arXiv, ICLR, CVPR, NeurIPS, ICCV \\
\Xhline{0.7pt}
\multirow{2}{*}{Paper} &
GPT-5 generated &
arXiv, D2L, Medium, CS231n &
arXiv, D2L, Medium, CS231n \\
\\
\Xhline{0.7pt}
\multirow{2}{*}{Art} &
WikiHow, EasyDrawing, YouTube, ArtForKidsHub &
WikiHow, EasyDrawing, YouTube, ArtForKidsHub &
WikiHow, EasyDrawing, YouTube, ArtForKidsHub \\
\Xhline{0.7pt}
\multirow{2}{*}{Life} &
WikiHow, YouTube &
WikiHow, YouTube &
WikiHow, YouTube \\
\\
\Xhline{0.7pt}
\multirow{2}{*}{Tech} &
WikiHow, Food52, YouTube &
WikiHow, Food52, YouTube &
WikiHow, Food52, YouTube \\
\\
\Xhline{0.7pt}
\multirow{2}{*}{Exercise} &
WikiHow, Healthline, Men's Health, YouTube &
WikiHow, Healthline, Men's Health, YouTube &
WikiHow, Healthline, Men's Health, YouTube \\
\shline
\end{tabular}
\caption{\paragraph{Per-task breakdown of question, reference image, and reference text sources} The table shows the diverse origins of data used for different tasks in \benchmarkname{}.}
\label{tab:detailed_task_info_split}
\end{table}

\section{Evaluation Details}
\label{sec:prompt}
\begin{figure}[h]
\centering
\includegraphics[width=\linewidth]{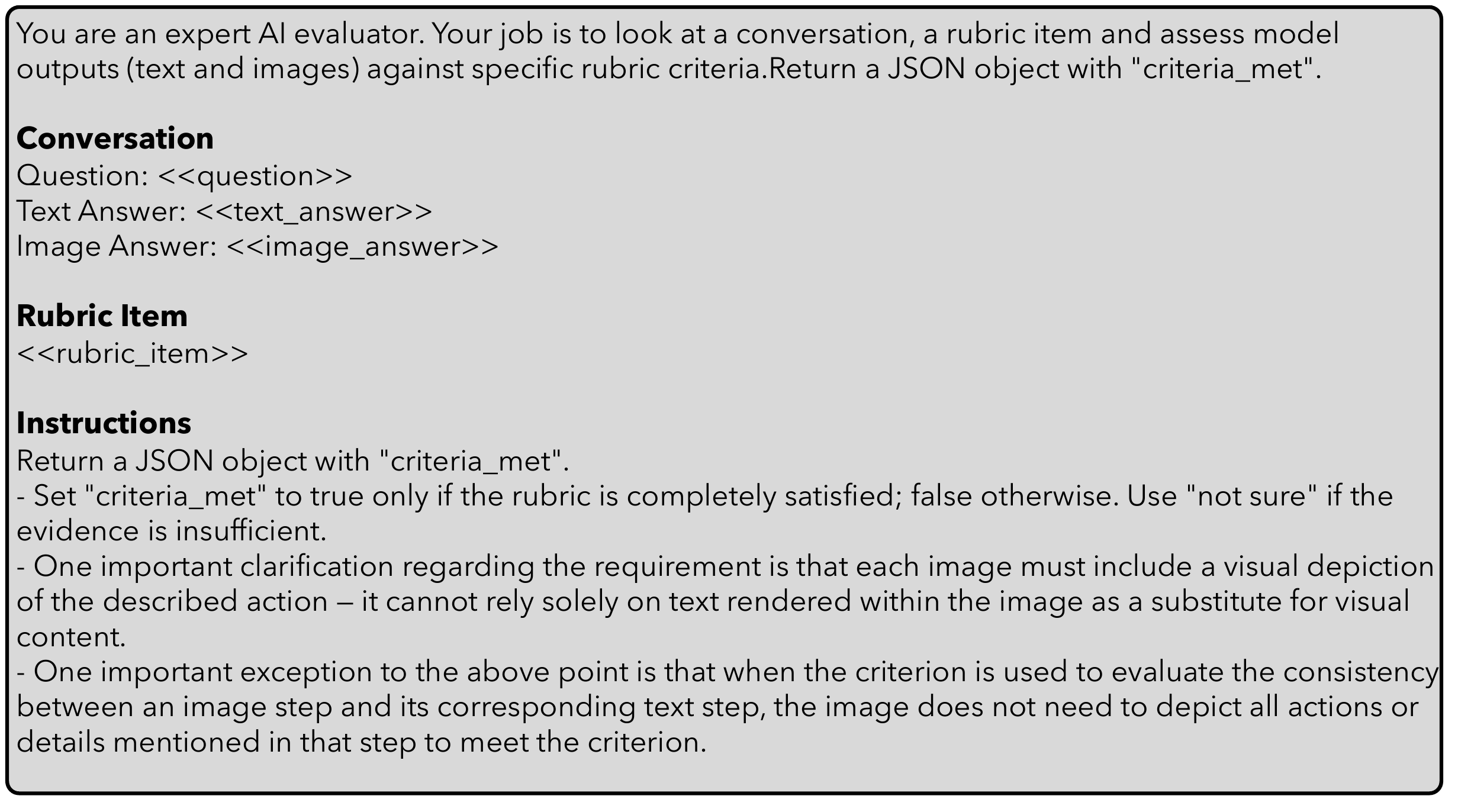}
\caption{\paragraph{Evaluation prompt used for Gemini-2.5-Pro as the judge model} It shows how a single rubric item is applied to evaluate model responses and produce structured judgments.}
\label{fig:evaluation_prompt}
\end{figure}

\figref{fig:evaluation_prompt} shows the prompt used to grade model responses (\ie, image + text) with an MLLM judge (\eg, Gemini-2.5-Pro) based on a specific rubric item during evaluation.

\clearpage
\newpage
\section{Responses from More Models}
~\label{sec:more_responses}

\begin{figure}[h]
\centering
\includegraphics[width=\linewidth]{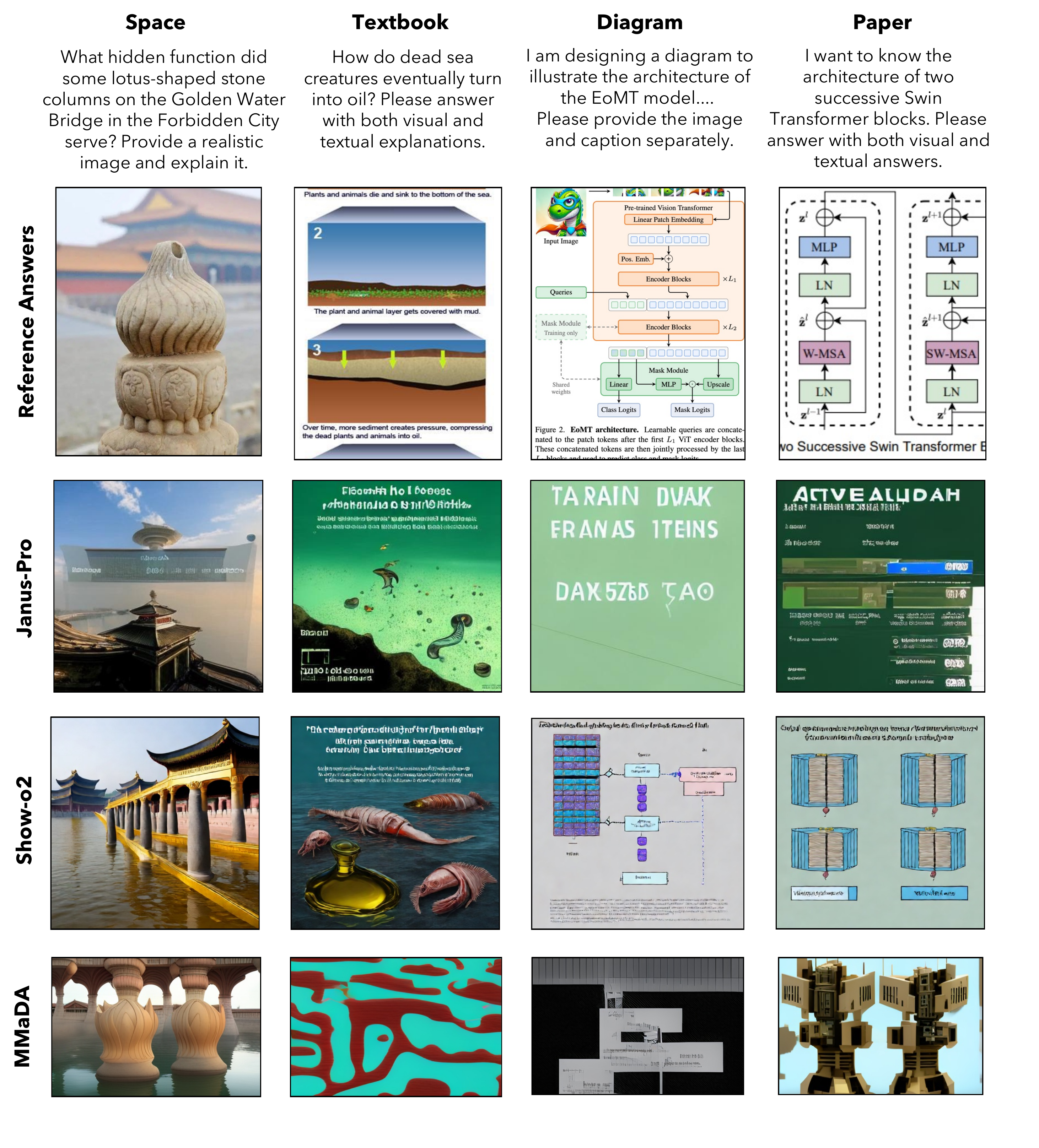}
\caption{\paragraph{Model-generated images for closed-ended tasks in \benchmarkname{}} We visualize images generated by Janus-Pro, Show-o2, and MMaDA.}
\label{fig:example1}
\end{figure}

\begin{figure}[h]
\centering
\includegraphics[width=\linewidth]{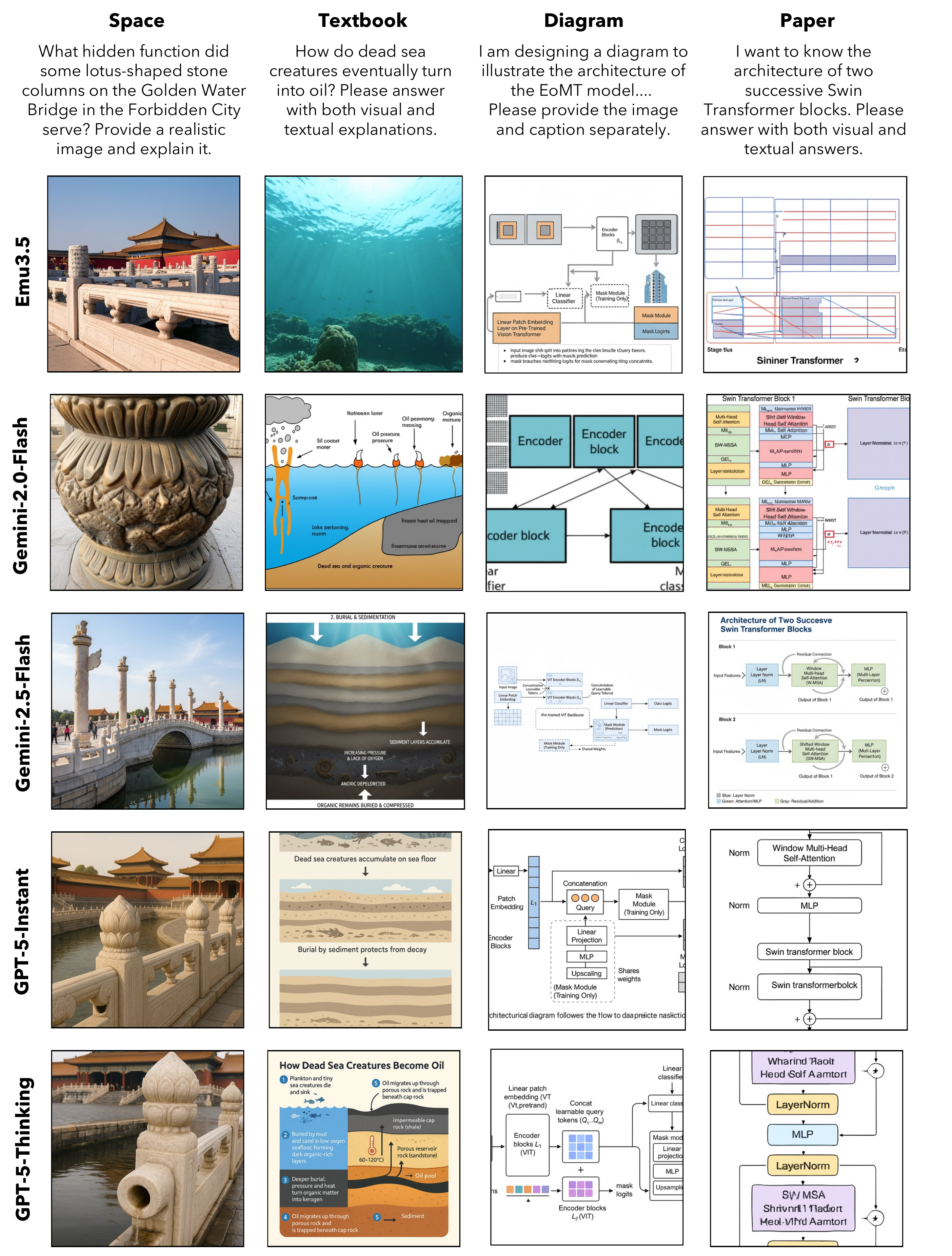}
\caption{\paragraph{Model-generated images for closed-ended tasks in \benchmarkname{}} We visualize images generated by Emu3.5, Gemini-2.0-Flash, Gemini-2.5-Flash, GPT-5-Instant, and GPT-5-Thinking.}
\label{fig:example2}
\end{figure}

\begin{figure}[t]
\centering
\includegraphics[width=\linewidth]{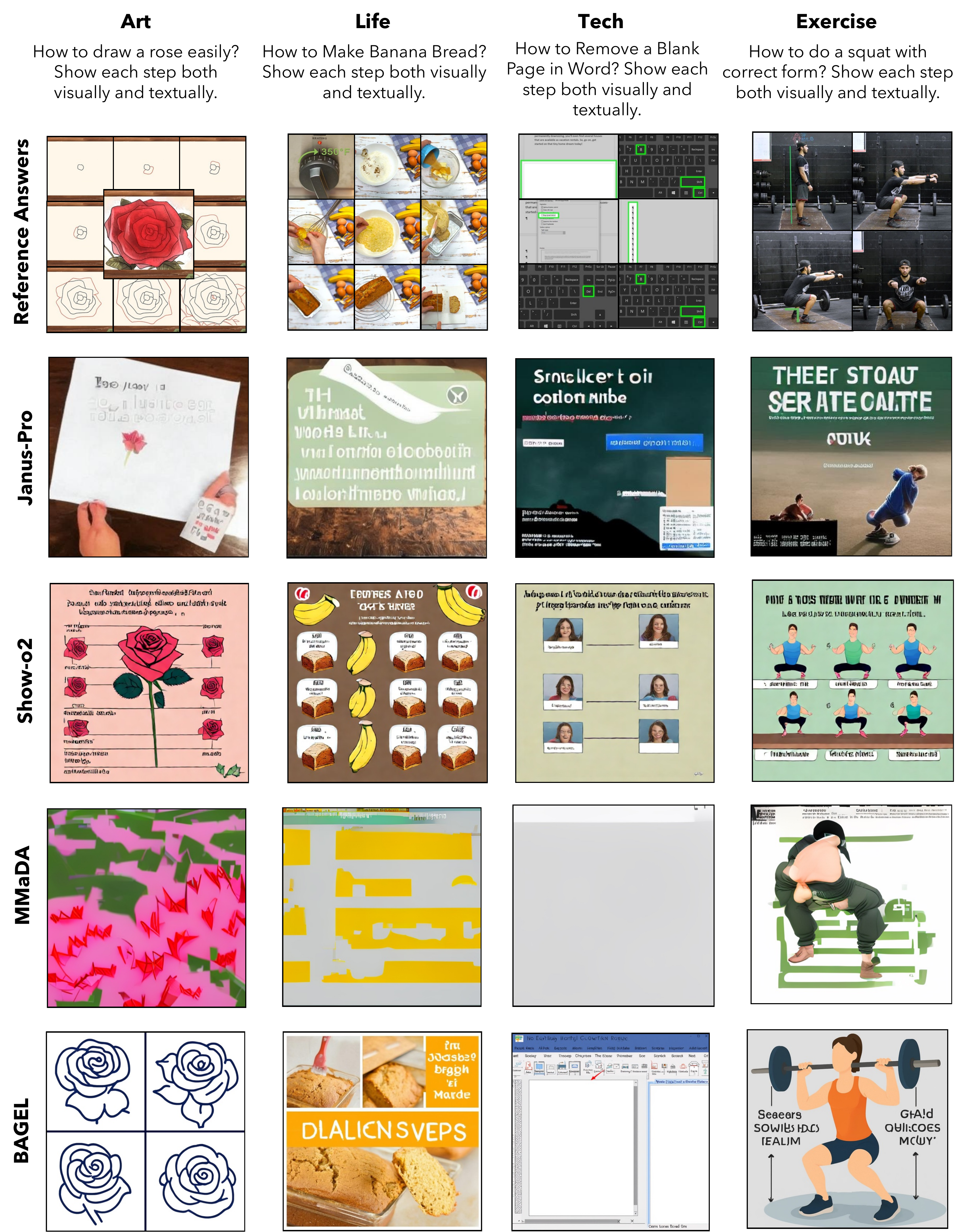}
\caption{\paragraph{Model-generated images for open-ended tasks in \benchmarkname{}} We prompt Janus-Pro, Show-o2, MMaDA, and BAGEL to produce step-by-step visual guides for each task.}
\label{fig:example3}
\end{figure}

\begin{figure}[t]
\centering
\includegraphics[width=\linewidth]{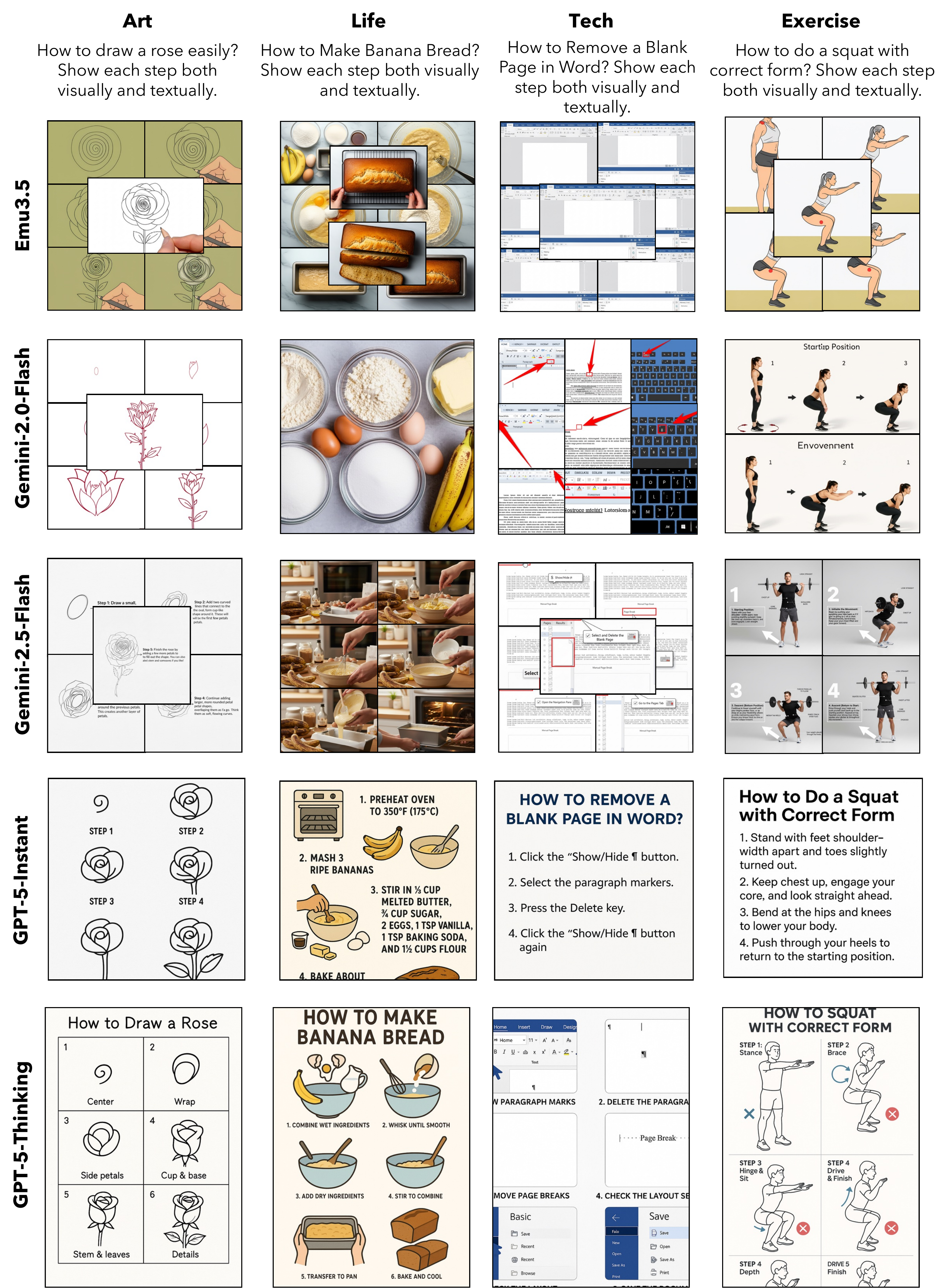}
\caption{\paragraph{Model-generated images for open-ended tasks in \benchmarkname{}} We prompt Emu3.5, Gemini-2.0-Flash, Gemini-2.5-Flash, GPT-5-Instant, and GPT-5-Thinking to produce step-by-step visual guides for each task.}
\label{fig:example4}
\end{figure}

Figures~\hyperref[fig:example1]{\textcolor{link}{\ref*{fig:example1}}},~\hyperref[fig:example2]{\textcolor{link}{\ref*{fig:example2}}},~\hyperref[fig:example3]{\textcolor{link}{\ref*{fig:example3}}}, and~\hyperref[fig:example4]{\textcolor{link}{\ref*{fig:example4}}} visualize additional examples of images generated by different models.

\end{document}